\documentclass[10pt, conference, letterpaper]{IEEEtran}
\usepackage{cite}
\usepackage{amsmath,amssymb,amsfonts}
\usepackage{graphicx}
\usepackage{textcomp}
\usepackage{xcolor}

\usepackage{xspace}
\usepackage[hidelinks]{hyperref}
\usepackage{cleveref}
\usepackage{booktabs}
\usepackage{caption}
\usepackage{subcaption}
\usepackage{algorithm}
\usepackage{algpseudocode}
\usepackage{multirow}

\usepackage[]{mdframed}

\def\BibTeX{{\rm B\kern-.05em{\sc i\kern-.025em b}\kern-.08em
    T\kern-.1667em\lower.7ex\hbox{E}\kern-.125emX}}

\newcommand{\myname}{FedFetch\xspace}
\algnewcommand{\LineComment}[1]{\State \(\triangleright\) #1}
\algnewcommand{\LineCommentBlue}[1]{\State \textcolor{blue}{\(\triangleright\) #1}}

\usepackage{tikz}
\usepackage{textcomp}
\usepackage{hyperref}
\usepackage{lipsum}

\newcommand\copyrighttext{%
  \footnotesize \textcopyright 2025 IEEE. Personal use of this material is permitted.
  Permission from IEEE must be obtained for all other uses, in any current or future
  media, including reprinting/republishing this material for advertising or promotional
  purposes, creating new collective works, for resale or redistribution to servers or
  lists, or reuse of any copyrighted component of this work in other works.
  DOI: \href{<http://tex.stackexchange.com>}{TBD}}
\newcommand\copyrightnotice{%
\begin{tikzpicture}[remember picture,overlay]
\node[anchor=south,yshift=10pt] at (current page.south) {\fbox{\parbox{\dimexpr\textwidth-\fboxsep-\fboxrule\relax}{\copyrighttext}}};
\end{tikzpicture}%
}

\newcommand\authorversiontext{%
  \footnotesize \textcopyright This is the author’s accepted version of the article. The final version published by IEEE is Q. Yan, A. Liu, S. He, M. Lécuyer and I. Beschastnikh, “FedFetch: Faster Federated Learning With Adaptive Downstream Prefetching”, IEEE INFOCOM 2025 - IEEE International Conference on Computer Communications, doi: TBD.}
\newcommand\authorversionnotice{%
\begin{tikzpicture}[remember picture,overlay]
\node[anchor=north,yshift=0pt] at (current page.north) {\fbox{\parbox{\dimexpr\textwidth-\fboxsep-\fboxrule\relax}{\authorversiontext}}};
\end{tikzpicture}%
}

\makeatletter
\newcommand{\algmargin}{\the\ALG@thistlm}
\makeatother
\algnewcommand{\ParState}[1]{\State%
  \parbox[t]{\dimexpr\linewidth-\algmargin}{\strut #1\strut}}

\begin{document}
\title{
FedFetch: Faster Federated Learning with Adaptive Downstream Prefetching
}

\author{\IEEEauthorblockN{Qifan Yan\IEEEauthorrefmark{1}, Andrew Liu\IEEEauthorrefmark{1}, Shiqi He\IEEEauthorrefmark{2}, Mathias Lécuyer\IEEEauthorrefmark{1} and Ivan Beschastnikh\IEEEauthorrefmark{1}}
\IEEEauthorblockA{\IEEEauthorrefmark{1}Department of Computer Science, University of British Columbia, Canada\\
Email: ericy676@student.ubc.ca, yul02@student.ubc.ca, mathias.lecuyer@ubc.ca, bestchai@cs.ubc.ca}
\IEEEauthorblockA{\IEEEauthorrefmark{2}Computer Science and Engineering, University of Michigan, USA\\
Email: shiqihe@umich.edu}
}

\maketitle

\begin{abstract}
Federated learning (FL) is a machine learning paradigm that facilitates massively distributed model training with end-user data on edge devices directed by a central server.
However, the large number of heterogeneous clients in FL deployments leads to a communication bottleneck between the server and the clients. This bottleneck is made worse by straggling clients, any one of which will further slow down training. To tackle these challenges, researchers have proposed techniques like client sampling and update compression. These techniques work well in isolation but combine poorly in the downstream, server-to-client direction. This is because unselected clients have outdated local model states and need to synchronize these states with the server first. 

We introduce \myname, a strategy to mitigate the download time overhead caused by combining client sampling and compression techniques. \myname achieves this with an efficient prefetch schedule for clients to prefetch model states multiple rounds before a stated training round.
We empirically show that adding \myname to communication efficient FL techniques reduces end-to-end training time by 1.26$\times$ and download time by 4.49$\times$ across compression techniques with heterogeneous client settings. Our implementation is available at \url{https://github.com/DistributedML/FedFetch}

\end{abstract}


\authorversionnotice
\copyrightnotice 

\section{Introduction}

In Federated learning (FL) a set of distributed clients collaboratively train an ML model with the help of a central parameter server~\cite{mcmahan2017FedAvg, kairouz2021advances}. The clients train with their local data which they never share publicly; instead, clients send their local models or the corresponding gradients to the server for aggregation. This feature enables FL to source data from edge clients without needing to pool data into a single location.

Our work focuses on the cross-device FL setting in which a large number of heterogeneous edge clients train a model (e.g., mobile phones, laptops, IoT devices)~\cite{kairouz2021advances}. Following previous works, we categorize heterogeneity into system and statistical heterogeneity~\cite{oort-osdi21, luo2022adaptiveClientSampling}. System heterogeneity refers to the different network bandwidth capacity, compute capacity, and device availability of clients. Statistical heterogeneity focuses on the dissimilarity of training data characteristics, like the number of samples, presence of labels, quality of examples, etc. In general, the client training data is not independently and identically distributed (non-iid).

Due to a large number of heterogeneous clients in cross-device FL,  the transfer of updates between the server and clients consumes a significant amount of time and bandwidth, especially in the downstream, server-to-client, direction. For example,  Google's production FL system with over  600 clients selected for training every round, with peak server traffic of around 600 MB/s for downstream and 200 MB/s for upstream updates~\cite{bonawitz2019FLAtScale}. Furthermore, straggling clients with low bandwidth or compute capacity inflate the training process.
We  consider two types of approaches to reduce communication costs in terms of both time and bandwidth: client sampling~\cite{mcmahan2017FedAvg, he2023gluefl, oort-osdi21, PyramidFL-Li2022, luo2022adaptiveClientSampling, chen2021communication, chen2022optimalClientSampling, nishio2019clientSelectionSystem, wu2023anchor} and update compression. The latter can be further categorized into masking/sparsification~\cite{he2023gluefl, sattler2019STC, SparseSGD2018Stich, wu2024fiarse, chen2021APF}, quantization~\cite{alistarh2017qsgd, seide2014-1bit, EDEN-pmlr-v162-vargaftik22a, DoCoFL-pmlr-v202-dorfman23a, LFL-amiri2020, zheng2020designLayeredDownQuant, wu2022sign}, low-rank decomposition~\cite{vogels2019powersgd}, and sketching~\cite{rothchild2020fetchsgd}. 

To save bandwidth and training time, cross-device FL deployments rely on a combination of client sampling and compression. 
However, recent work highlighted that the time and bandwidth improvements that client sampling and compression bring diminish significantly when they are combined in the downstream direction~\cite{he2023gluefl, DoCoFL-pmlr-v202-dorfman23a}. 
For instance, He et al.~\cite{he2023gluefl} found that a naive combination of client sampling with masking is ineffective in the downstream direction. This is because of client model \emph{staleness}, which is when a client model is not up to date with the server's model due to clients not participating in every training round. Model staleness also comes up with non-masking techniques like quantization and low-rank decomposition. Most work on quantizing downstream model updates either assumes full participation to circumvent client model staleness~\cite{LFL-amiri2020} or full model synchronization before every FL round~\cite{alistarh2017qsgd, vogels2019powersgd, EDEN-pmlr-v162-vargaftik22a}.
Consequently, the need to synchronize models slows down FL deployments. Client heterogeneity further exacerbates this issue, as clients with weaker connectivity will inflate the time to download large downstream updates.

We present \textbf{\myname}, a general FL method to address the time delay related to synchronizing stale client models caused by client sampling and compression. 
A standard FL system has a single \emph{Train} phase in every round, which includes client synchronizing a server model, performing local training, and sending  results for server aggregation.
\myname introduces two new phases that come before the \emph{Train} phase: \emph{Prepare} and \emph{Prefetch}.

During the \emph{Prepare} phase, the server presamples the clients that will run $R$ rounds in the future. For each sampled client, the server will create a customized download schedule for the client, depending on knowledge about the client's bandwidth profile. In the \emph{Prefetch} phase, the clients download the latest global model updates according to their schedules.

\begin{mdframed}
In summary, \myname shifts client downstream bandwidth usage from the \emph{Train} phase to the \emph{Prefetch} phase. This reduces end-to-end training time.
\end{mdframed}

Overall, we make the following contributions:

\begin{itemize}
    \item We characterize the deficiencies of naively combining client selection and compression in downstream communication under heterogeneous cross-device FL conditions. We observe that clients need to synchronize a larger update for each round missed due to not being selected.

    \item We introduce \myname, a general prefetching framework for cross-device FL. \myname reduces client local model staleness during client sampling to shorten end-to-end and download time in cross-device FL by 1.26$\times$ and 4.49$\times$ with an 12\% extra bandwidth cost.
    
    \item We evaluate \myname's compatibility and ease of integration with representative client sampling \cite{mcmahan2017FedAvg, he2023gluefl} and compression techniques~\cite{he2023gluefl, sattler2019STC, LFL-amiri2020, alistarh2017qsgd, EDEN-pmlr-v162-vargaftik22a, DoCoFL-pmlr-v202-dorfman23a, vogels2019powersgd} in environments with system and statistical heterogeneity. We find that \myname consistently decreases the downstream state synchronization time for every method.
\end{itemize}

\section{Background and Motivation}

\subsection{Cross-device FL Characteristics}

Cross-device FL is characterized by a high level of system heterogeneity. The differences between clients arise from various sources, such as the type of device, service provider, geographical location etc. In standard FL designs, an FL communication round concludes only when the stragglers (i.e., slowest clients) finish. These stragglers harm performance since client bandwidths vary by orders of magnitude.  

\begin{figure}[t!]
     \centering
     \begin{subfigure}[t]{0.49\linewidth}
        \centering
        \includegraphics[width=1.0\linewidth]{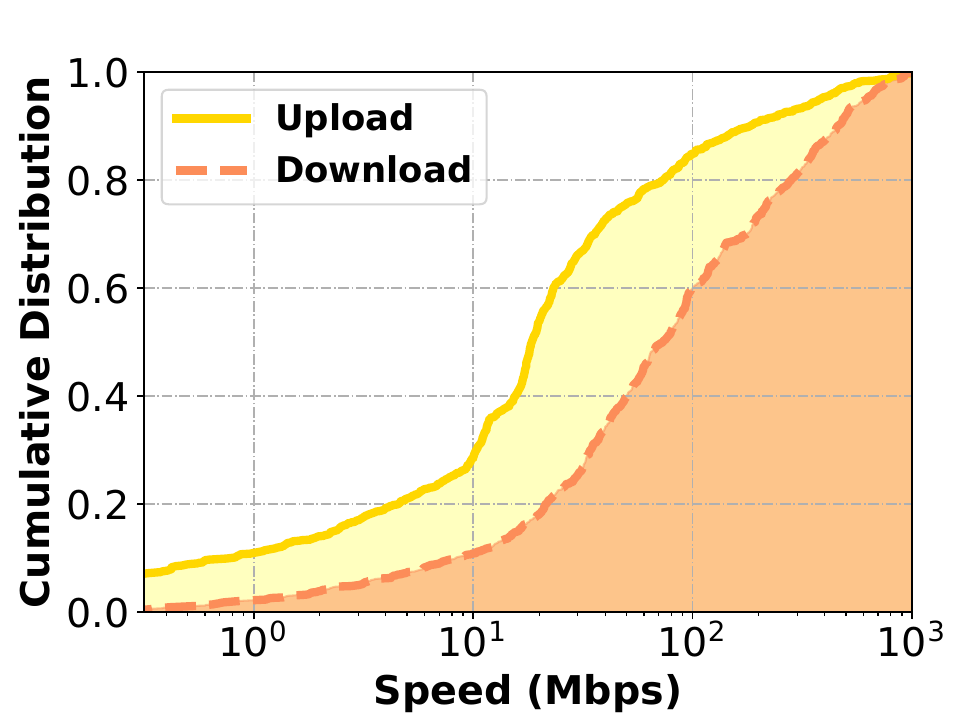}
        \caption{CDF of edge device download and upload bandwidth distribution for North America in Jan 2024~\cite{mlab}.}
        \label{fig:speed-test}
     \end{subfigure}
     \hfill
     \begin{subfigure}[t]{0.49\linewidth}
         \centering
         \includegraphics[width=1.0\linewidth]{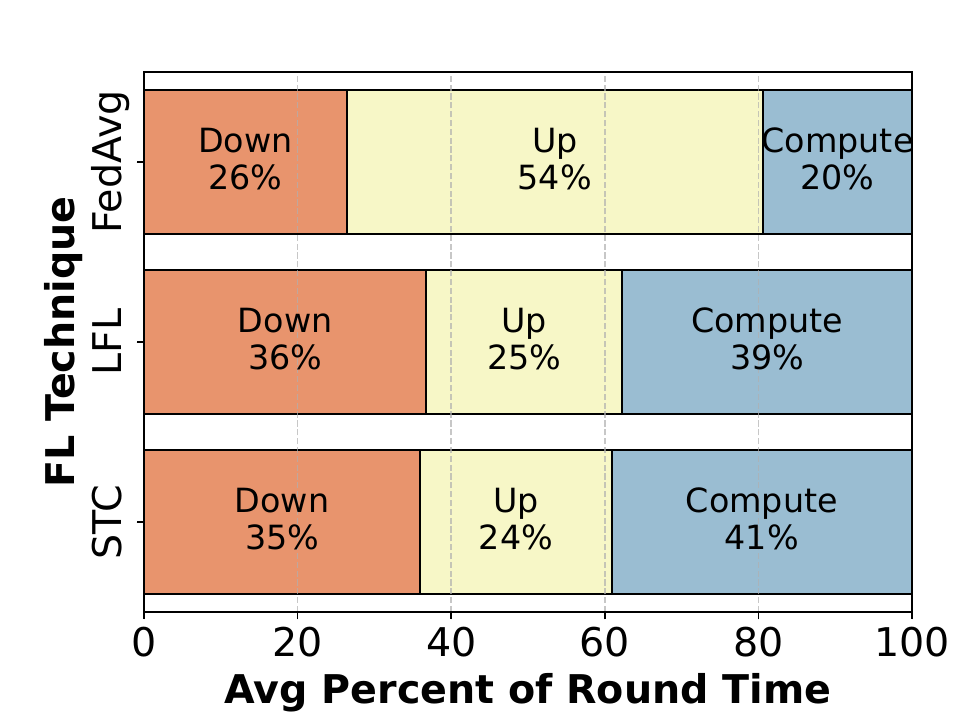}
         \caption{Percentage breakdown of an FL round in terms of Download (Down), Upload (Up), and Compute time.}
         \label{fig:fl-time-ratio}
     \end{subfigure}
    \caption{Cross-device FL Characteristics.}
\end{figure}

To illustrate the effect of system heterogeneity, we plotted the download and upload speeds of edge clients, such as mobile phones and personal computers,
in Figure~\ref{fig:speed-test}. The figure uses data from Measurement Lab's NDT speed test dataset for N.America in Jan 2024~\cite{mlab}. Note that roughly 5\% of clients have download speeds of less than 4 Mbps, which is about 25 times slower than the median speed of 81.29 Mbps. 

\Cref{fig:fl-time-ratio} shows results from an experiment with the distribution in Figure~\ref{fig:speed-test}. It records the breakdown of an FL round in terms of every round's average download, upload, and compute times. We ran these experiments with the FEMNIST dataset using the setup in~\Cref{sec:experiment-setup}. This figure shows that communication can become a major bottleneck in cross-device FL settings with FedAvg~\cite{mcmahan2017FedAvg}, consuming nearly 80\% of the total time in a round. Moreover, upstream communication is the most time-consuming because client upload speeds are typically slower than download speeds (\Cref{fig:speed-test}). This communication overhead motivates the need for communication reduction techniques such as masking with STC~\cite{sattler2019STC} or quantization with LFL~\cite{LFL-amiri2020}. However, \Cref{fig:fl-time-ratio} also shows that while these optimizations reduces upstream communication time, they fail to reduce downstream communication. We explore why this is the case in~\Cref{sec:compression-bg}.

Availability of clients is another key issue in cross-device FL. Clients may spontaneously go offline for various reasons, such as the device running out of power or losing connectivity. In production systems, around 10\% of clients may drop out during each round~\cite{Yang2024FLASH, bonawitz2019FLAtScale}. These device failures slow down model convergence, extend training time, and waste hardware resources.  
Practical cross-device federated learning systems use an over-commitment ($OC$) mechanism \cite{fedscale-icml22, he2023gluefl, bonawitz2019FLAtScale}, which selects extra clients to participate in each round to mitigate unavailable and slow clients.

\subsection{Client Sampling}
\label{sec:sampling-bg}
A closely studied approach to reduce communication cost is client sampling, which, in its basic form, uniformly samples a fraction of the total number of clients to participate in each round of FL~\cite{he2023gluefl, mcmahan2017FedAvg}. Various sampling techniques have been proposed to further improve time/bandwidth-to-accuracy performance~\cite{fu2023clientSelectionSurvey}, ranging from choosing clients with better bandwidth/computation capacity \cite{nishio2019clientSelectionSystem} or those more likely to improve convergence~\cite{chen2021communication, chen2022optimalClientSampling, wu2023anchor} or a combination of the two~\cite{oort-osdi21, PyramidFL-Li2022}. Sampling reduces downstream (server to client) and upstream (client to server) bandwidth because of fewer participating clients per round.

Unfortunately, client sampling leads to client-side model staleness. Specifically, clients may now have to wait many rounds before being sampled or resampled. For instance, in simple random sampling with uniform probabilities, a client is expected to be resampled every $N/K$ rounds~\cite{he2023gluefl}. In effect, client sampling causes the client's local model to become stale relative to the latest server model.

\subsection{Compression} 
\label{sec:compression-bg}

In this paper, we focus on combining client sampling with three representative compression techniques: \emph{masking}, \emph{quantization}, and \emph{low-rank decomposition}. 

\emph{\underline{Masking}} techniques such as sparsification~\cite{sattler2019STC, he2023gluefl, SparseSGD2018Stich, wu2024fiarse} and parameter freezing~\cite{chen2021APF} are commonly used for compressing model updates. Specifically, masks are locations of parameters in a model that are transferred along with the parameter value. Usually, the chosen locations correspond to the most useful values in an update. For example, in $TopK$ sparsification, the top $K$\%  parameters, ordered by their absolute value, are transmitted along with their position information~\cite{SparseSGD2018Stich}. 

\emph{\underline{Quantization}} is another family of techniques to reduce communication costs in FL~\cite{alistarh2017qsgd, seide2014-1bit, EDEN-pmlr-v162-vargaftik22a, DoCoFL-pmlr-v202-dorfman23a, LFL-amiri2020, zheng2020designLayeredDownQuant, wu2022sign}. Quantization reduces the encoding precision of the model update to reduce communication, by casting a higher-bit representation of the update into a smaller lower-bit representation.

\emph{\underline{Low-rank decomposition}} techniques \cite{vogels2019powersgd} is the third category of compression techniques we consider in \myname. Low-rank methods rely on factorizing an input matrix $M\in \mathbb{R}^{a,b}$ into two low-rank matrices $P \in \mathbb{R}^{a \times r}$ and $Q \in \mathbb{R}^{r \times b}$ where $r \ll \min(a, b)$. The resulting matrices are significantly smaller than the input matrix.

These compression techniques are straightforward to apply in the upstream direction. Yet, the downstream direction is troublesome with client sampling. The reason is that clients selected for training should have the most up-to-date model parameters. In the case of masking, due to the changing server-side mask, an increasing proportion of parameters will have changed with every round passed since the last time a client participated. For clients equipped with a stale local model, synchronization often leads to downloading the entire model at the start of their \emph{Train} phase, instead of a smaller masked update. This nullifies speed and bandwidth improvements in the downstream update synchronization. Downstream compression with quantization, and low-rank decomposition are also difficult when a client's model is out-of-date.

\subsection{Quantifying Staleness}
\label{sec:quantify-staleness}

\begin{figure}
     \centering
     \begin{subfigure}[t]{0.49\linewidth}
         \centering
         \includegraphics[width=\linewidth]{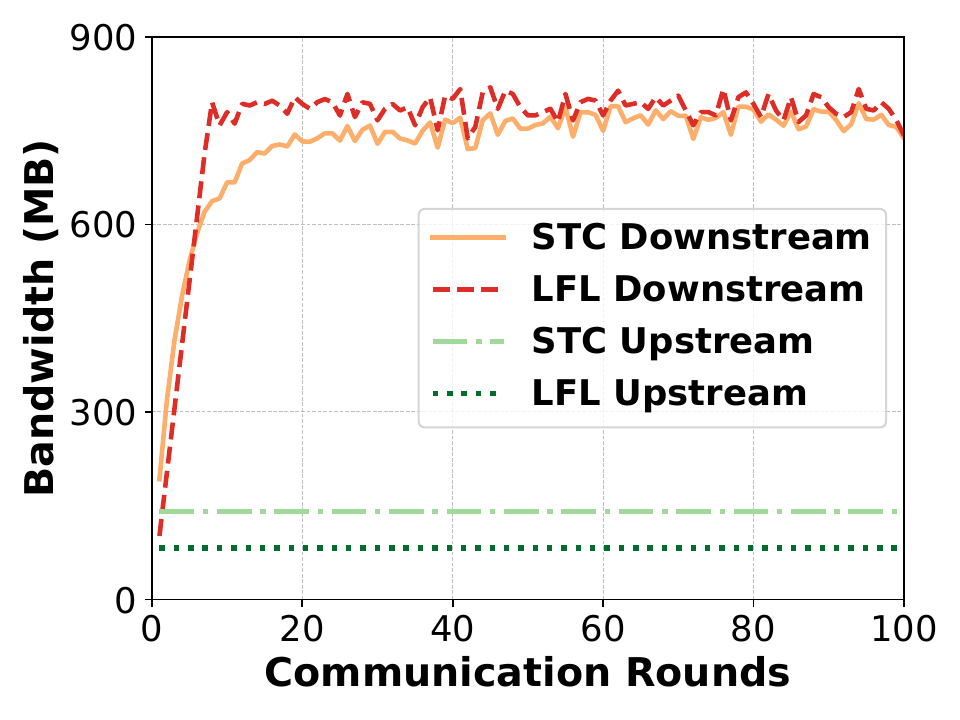}
         \caption{Download and upload amount per round.}
         \label{fig:staleness-dl-ul}
     \end{subfigure}
     \hfill
     \begin{subfigure}[t]{0.49\linewidth}
         \centering
         \includegraphics[width=\linewidth]{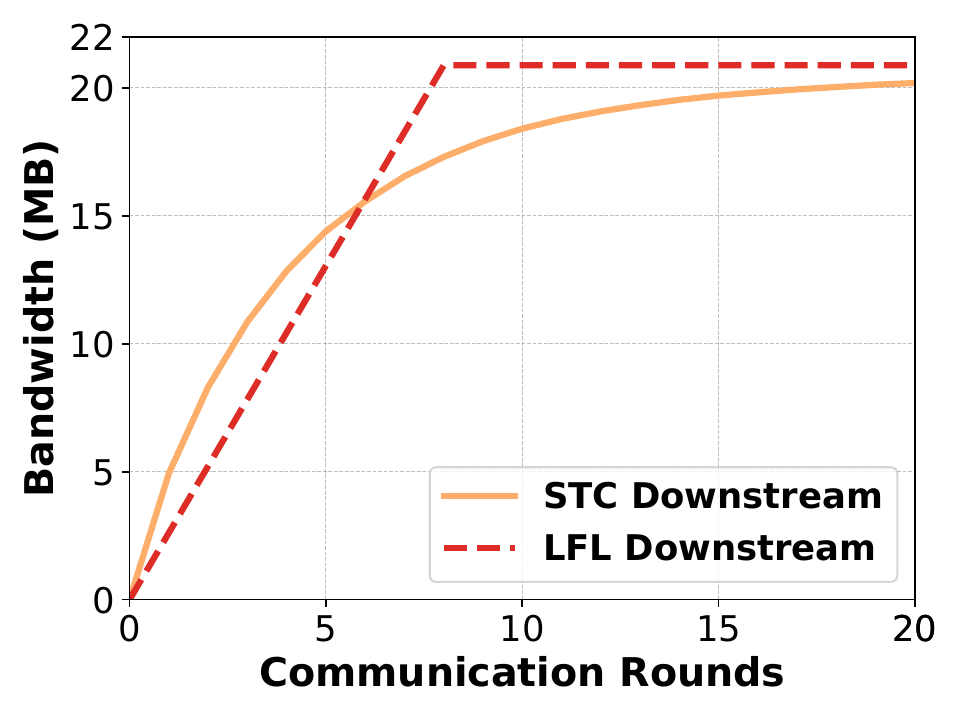}
         \caption{Average download sizes for clients who are resampled after a number of rounds.}
         \label{fig:staleness-dl}
     \end{subfigure}
    \caption{
    Effect of combining client sampling and compression.
    \label{fig:staleness}
    }
\end{figure}

\begin{figure*}
    \centering
    \includegraphics[width=.8\linewidth]{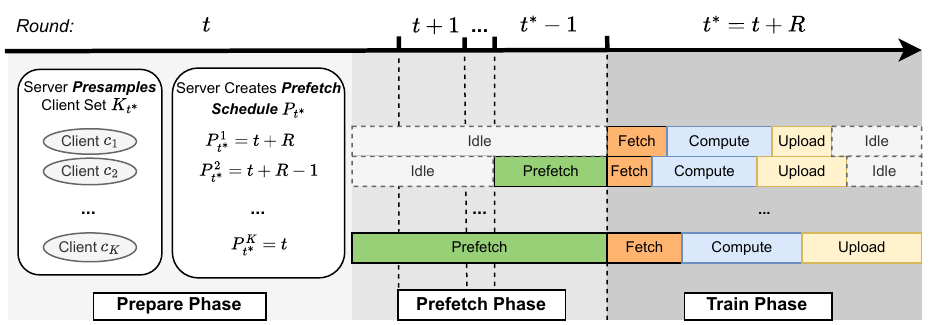}
    \caption{\myname Design. The goal of \myname is to minimize the amount of time clients spend on model download during their \emph{Train} phase (\emph{``Fetch"} in orange in the diagram). \myname introduces two new phases: \emph{Prepare} and \emph{Prefetch}. During \emph{Prepare}, clients are presampled by the server and provided with a prefetch schedule. During \emph{Prefetch}, each client prefetches model state (\emph{``Prefetch"} in green in the diagram) from the server before their \emph{Train} phase starts.}
    \label{fig:overall-design}
\end{figure*}

To better understand the staleness problem, we conduct 
an experiment using simple random client sampling with STC\cite{sattler2019STC}, a masking method; and, LFL\cite{LFL-amiri2020}, a quantization technique using default settings from \Cref{sec:parameters}. 

\Cref{fig:staleness-dl-ul} plots communication volume in downstream and upstream directions for FL using STC with a sparsification ratio of 0.2 and LFL with 4 bits. 
\Cref{fig:staleness-dl-ul} shows that despite the large and consistent savings in the upstream direction, there is little downstream savings past the first 20 rounds. Most clients sampled every round have not participated in training recently. Consequently, these clients must download large updates (possibly the entire model) to catch up.

In \Cref{fig:staleness-dl}, we plot how much content a resampled client needs to download after \emph{not} being selected for a variable number of rounds, for the two methods. The key observation is that with growing local model staleness, a client needs to download an increasingly large update. This update approaches the full model size after 15 rounds for STC and 8 rounds for LFL.

\begin{table}
  \caption{Summary of notations.}
  \label{tab:notations}
\begin{center}
\begin{tabular}{cl}
    \toprule
    \textbf{Symbol} & \textbf{Definition}\\
    \midrule
    $t, T$ & index and total number of \textit{rounds}\\
    $i, N, \mathcal{N}$ & index, total number, set of \textit{clients}\\
    $K, \mathcal{K}_{t}$ & number and set of \textit{clients} sampled at $t$\\
    
    $w_{t}, \hat{w}_{t}^i$ & server and client \textit{model} at the start of $t$\\
    $\nu_{i}$ & aggregation weight of client $i$\\
    $\hat{\Delta}_t^i, \Delta$ & client $i$'s and aggregated server \textit{update} at $t$\\
    $R$ & max number of rounds available for prefetch
    \\
    $P_{t}^i$ & prefetch schedule for round $t$\\
    $C_{dl}, C_{ul}$ & downlink and uplink \textit{compressor}\\
    $\delta_{t_1,t_2}$ & accumulated server updates from $t_1$ to $t_2$\\
    $d_t, D_t$ & true and estimated round duration at $t$\\
    $BW_{dl}^i$ & downlink bandwidth for $i$\\
    $OC$ & over-commitment\\
    \bottomrule
\end{tabular}
\end{center}
\end{table}

\subsection{Prefetching in FL} 
Prior research has considered the importance of clients synchronizing the most recent model state~\cite{sattler2019STC, LFL-amiri2020, he2023gluefl} or a relatively recent state~\cite{DoCoFL-pmlr-v202-dorfman23a} before receiving compressed downstream updates. With the exception of~\cite{DoCoFL-pmlr-v202-dorfman23a}, these methods use simple designs with state synchronization at the start of their training round ($R=0$) or one prior round ($R=1$). 

DoCoFL~\cite{DoCoFL-pmlr-v202-dorfman23a} assigns clients to start their \emph{Train} phase with a fixed and predetermined time window. For dealing with heterogeneous client bandwidths, they proposed but did not seem to evaluate, two separate time windows: one for strongly-connected clients and another for weakly-connected clients. This approach is unrealistic for cross-device FL environments where round durations vary~\cite{bonawitz2019FLAtScale, fedscale-icml22}. 

In the simple forms of state synchronization above, all clients scheduled for the same round of future training will prefetch a fixed update. However, 
as we demonstrate in \Cref{sec:quantify-staleness}, clients with knowledge of the more recent global models can download \emph{smaller} updates. Simple strategies fail to take advantage of this opportunity. 

\section{\myname Design}
We introduce \underline{prefetching} as a new dimension for time-to-accuracy optimization in cross-device FL which we realize in \emph{\myname}. We now explain its design.

\Cref{fig:overall-design} shows how \emph{\myname} introduces two new phases to FL: \emph{Prepare} and \emph{Prefetch}. The server-side \emph{Prepare} phase pre-determines which clients will be selected and when they will start prefetching. The client-side \emph{Prefetch} phase features clients prefetching model states before their scheduled training round. Both phases are controlled by the hyperparameter $R$, the number of rounds a client will be presampled in advance and the maximum number of rounds available for a client to prefetch. If $R=0$, then \myname is equivalent to the standard synchronous FL algorithm. In \myname, we are interested in cases where $R \geq 1$.

\subsection{Prepare phase}
\label{sec:preparation}
The \emph{Prepare} phase has two objectives. First, the server creates a maximum prefetch time budget for clients to prefetch the latest global model before the start of their training round with presampling. Second, the server adjusts the prefetch budget for each client to reduce total downlink update sizes while minimizing download time with prefetch scheduling. 

\subsubsection{Presampling}
\label{sec:presampling}
In presampling, the server samples clients who will participate in a future training round. In general, at round $t$, the server will generate $K_{t^*}$, where $t^*=t+R$, the set of $K$ clients scheduled to commence training in round $t^*$. For simplicity, \Cref{alg:FedFetch} only shows clients that are presampled on round $t = 1$ or later who will start their \emph{Train} phase on round $R+1$ or later. \myname conducts no presampling and prefetching for clients with \emph{Train} phase in rounds $1, \dots, R$.

Critically, presampling is compatible with most client sampling strategies. 
We need to consider what changes across FL training rounds to understand why this is the case. Typically, the system profile, such as bandwidth and compute capacities,
are stable during an FL round, which takes only a few minutes. This means sampling clients a few rounds beforehand based on the client's system profile~\cite{nishio2019clientSelectionSystem} is sufficient. But, a client's statistical utility is  determined through profiling its local data or local model against information on the server~\cite{fu2023clientSelectionSurvey}. This profiling typically happens when the client last participated in training. The latter approach incurs additional communication overhead associated with synchronizing recent models just for profiling purposes~\cite{oort-osdi21}. As clients will take many rounds to be resampled anyway, we argue that predetermining selected clients only a few rounds in advance has little impact on determining a client's statistical utility.%

\begin{algorithm}[tbt!]
\caption{\myname Prefetch Scheduler}\label{alg:prefetch-scheduling}
\begin{algorithmic}[1]
\Procedure{SchedulePrefetch}{Presampled clients set $\mathcal{K}_{t^*}$, current round $t^S$, \emph{Train} phase round $t^*$}
\State $t^P \gets t^S$
\State $T_{limit} \gets \infty$
\For{$t \gets t^S, \dots, t^*$}
    \ParState{$\mathcal{T}_t \gets \{T_t^i = \textsc{EstFetchTime}(i) \mid$  $i \in  \mathcal{K}_{t^*} \}$ }
    \LineCommentBlue{Find clients with \emph{Train} phase fetch time  $\leq T_{limit}$}
    \State $\mathcal{F}_r \gets \{i \mid i \in \mathcal{K}_{t^*}$ and $T_t^i \leq T_{limit} \}$
    \If{$|\mathcal{F}_r| = |\mathcal{K}_{t+R}|$}
        \State $t^P \gets t$
        \State $T_{limit} \gets 1/OC$ percentile of $\textsc{SortAsc}(\mathcal{T}_t)$ 
    \EndIf
    \State $\forall i \in \mathcal{F}_r, P^i \gets t$ \label{alg:update-schedule}
\EndFor
\State \Return $P_{t^*} = \{P^i \mid i \in \mathcal{K}_{t^*} \}$
\EndProcedure
\State
\Procedure{EstFetchTime}{Client $i$, prefetch schedule round $t$, base model round $t^P$ }\label{alg:est-fetch-time}
\State $D_{avg}\gets\textsc{ExpWeightedAvgRoundDuration}()$
\State $U \gets \overline{w_{t}}$ \textcolor{blue}{\Comment{Accumulated updates to download}}
\State $B \gets 0$ \textcolor{blue}{\Comment{Client's prefetch time budget}}
\ParState {$l \gets t$ \textcolor{blue}{\Comment{Client's model is in sync with server model $w_l$ once it finishes downloading everything in $U$}}}
\For{$j \gets t, \dots, t^*-1$}
    \State $B \gets \max(0, B + D_{avg} - U / BW_{dl}^i$)
    \If{$B > 0$}
        \State $U \gets \max(0, \delta_{l, j-1} - B \cdot BW_{dl}^i)$
        \State $l \gets j$
    \Else 
        \State $U \gets U - D_{avg} \cdot BW_{dl}^i$
    \EndIf
\EndFor
\State \Return $(U + \delta_{l,t^*-1}) / BW_{dl}^i$
\EndProcedure
\end{algorithmic}
\end{algorithm}

\subsubsection{Prefetch Scheduling}
\label{sec:preample-schedule-estimation}
After the server selects all the clients $K_{t^*}$ to start their \emph{Train} phase in a future round $t^*$, it will need to construct a prefetch schedule for every client. 

To see why \myname needs a prefetch scheduler and cannot simply assign a fixed prefetch round for every client, we can consider the following scenario: The system presamples two clients $c_1$ and $c_2$. The first client, $c_1$, is a straggler with low downstream bandwidth and need multiple rounds to finish prefetching all necessary updates. $c_2$ is a non-straggler with high downstream bandwidth. To optimize the download time in the \emph{Fetch} phase, the server would prefer the client to prefetch as much as possible with a larger time budget in the \emph{Prefetch} phase. Hence, more rounds (up to the maximum of $R$ round) should be allocated to $c_1$. However, as we increase the prefetch window, we also inevitably increase the overall size of the downstream update (see \Cref{sec:prefetch-content}). This is inefficient for faster clients like $c_2$ who do not need that much time for prefetching. Therefore, the design goal for the prefetch scheduler is to assign the latest possible prefetch schedule for every client while not negatively impacting the time savings associated with a choice of $R$.

\Cref{alg:prefetch-scheduling} describes \myname's prefetch scheduling algorithm. The  scheduler follows an iterative process of determining whether for the current prefetch start round $t$, a client $c_i$ can acquire its \emph{Prefetch} and \emph{Train} phase downstream updates before a time limit $T_{limit}$. If possible, \myname updates the client's prefetch schedule to $t$. Otherwise, the client keeps its original schedule. In the scenario where all clients can complete within the time limit $T_{limit}$, the scheduler will advance $t^P$, the minimum number of rounds required for every client to finish prefetching within the time limit. 
There are two under-specified elements in this design:
\begin{itemize}
    \item What should the time limit $T_{limit}$ be?
    \item How can we determine the time it takes for a client to finish downloading its Prefetch and \emph{Train} phase downstream updates in some future round?
\end{itemize}

To answer the first question, we define the time limit $T_{limit}$ as the time required for the slowest client to finish acquiring all its downstream updates at the start of the \emph{Fetch} phase. If over-commitment $OC > 1$, then $T_{limit}$ becomes the $1 / OC$ percentile of fetch times to account for the fact that only the $K (1/OC)$ client updates are aggregated every round\footnote{We considered adding a factor $\beta$, where $1+\beta < OC$ so that $T_{limit}$ is the $(1+\beta) / OC$ percentile fetch time. We evaluated different values of $\beta$ and found no difference in results as long as $1+\beta$ is not close to $OC$.}.

To answer the second question, we introduce the \textsc{EstFetchTime} function in \Cref{alg:prefetch-scheduling} line \ref{alg:est-fetch-time}. This function takes in a client $c_i$'s bandwidth profile $BW_{dl}^i$ and some prefetch round $t$. Similar to~\cite{nishio2019clientSelectionSystem}, the client will immediately provide their bandwidth profile $BW_{dl}^i$ upon being presampled. The function then estimates, for the given client, the amount of time it will take to download all the required downstream updates (see \Cref{sec:prefetch-content}). It does this by simulating \myname's \emph{Prefetch} phase (see \Cref{sec:prefetch}). We choose to estimate the fetch time instead of the total \emph{Train} phase time because, unlike bandwidth, compute speed is harder to profile and would add uncertainty. Moreover, we highlight two differences between time calculations in estimations for prefetch scheduling and in practice during the \emph{Prefetch} phase.

First, we do not know the duration of future rounds during prefetch scheduling. To address this challenge, we estimate the round duration $D_{t}$ with an exponential weighted moving average of prior round durations with $\alpha=0.125$. This is the standard approach for estimating round trip times in protocols like TCP~\cite{rfc6298}. Specifically, the estimated round duration for the current round $D_{t}$ equals the weighted sum of the true round duration for the previous round $d_{t-1}$ and the past estimated duration $D_{t-1}$.
\begin{equation}
    \label{equ:duration-estimate}
    D_{t} = \alpha \cdot d_t + (1 - \alpha) \cdot D_{t-1}
\end{equation}
This better captures trends in round duration throughout the day~\cite{fedscale-icml22, bonawitz2019FLAtScale}. In our evaluation we found that the choice of $\alpha$ had little effect on our results, possibly due to the over-commitment mechanism removing extreme stragglers which decreases the variance of round durations.

Second, we do not know the size of various compressed updates exactly for masking techniques because the mask changes unpredictably across rounds. However, the absolute size of the accumulated update after a certain number of rounds is relatively stable (see \Cref{sec:quantify-staleness}). With this in mind, the server can dynamically profile the size of different rounds of accumulated updates by recording and averaging the sizes of updates sent during the \emph{Prefetch} and \emph{Train} phases. 

These differences between the prefetch scheduling and the actual \emph{Prefetch} phase mean that the scheduler can only approximate the optimal prefetch schedule. Nevertheless, we will show that this approximation leads to significant bandwidth savings in \Cref{sec:naive-prefetch-analysis}.

\subsubsection{Downstream Updates}
\label{sec:prefetch-content}

We now detail the downstream updates prefetched and fetched by clients during the \emph{Prefetch} and \emph{Train} phases.
Consider some round $t^S$ where the server presamples clients to start their \emph{Prefetch} phase in round $t^P$ and their \emph{Train} phase in round $t^*$. It is important to note that individual clients start prefetching at some round $P^i \in \{t^P, \dots, t^*\}$. The values for $P^i$ and $t^P$ are dynamically decided in \Cref{sec:prefetch-process}. \myname requires clients to synchronize a base model $w_{t^P}$ and all the server updates $\Delta_{t^P}, \dots, \Delta_{t^*-1}$ before the start of $t^*$. With downstream compression, these server updates will be further compressed with the downstream compressor $C_{dl}$. 
For brevity, we represent the sum of these compressed updates from round $t_1$ to $t_2$ with $t_1 \leq t_2$ with \Cref{equ:server-delta}. If $t_1 > t_2$, then $\delta_{t_1, t_2} = 0$.
\begin{equation}
    \label{equ:server-delta}
    \delta_{t_1, t_2} = \sum_{j = t_1}^{t_2} C_{dl}(\Delta_j)
\end{equation}
We proceed to summarize the full downstream update below.
\begin{equation}
    \label{equ:all-downstream}
    w_{t^*} = 
    w_{t^P} 
    + \delta_{t^P, t^*-1} 
\end{equation}

In terms of the sizes of each term in \Cref{equ:all-downstream}, $w_{t^P}$ has the same size as a full model. Each server update $C_{dl}(\Delta_j)$ has the minimum possible update size. However, transferring the sum of updates could lead to a smaller size than transferring each update separately (see \Cref{sec:quantify-staleness}). For example, $\delta_{1,2} = C_{dl}(\Delta_1) + C_{dl}(\Delta_2)$ could be smaller than transferring $\delta_{1,1} = C_{dl}(\Delta_1)$ and $\delta_{2,2} = C_{dl}(\Delta_2)$ separately if the downstream compressor $C_{dl}$ is a masking method.

Similarly, we can also combine the base model with all the updates until the client-specific prefetch schedule $P^i$ in a single $\overline{w_{P^i}}$ instead of transmitting $w_{t^P}$ and the updates $\delta_{t_p, P^i-1} $. 
These size reductions resulting from combining updates are what make the prefetch stage efficient. 
As a result, the prefetched and fetched update for a client is the following. 
\begin{equation}
    \label{equ:downstream-components}
    w_{t^*} = \overline{w_{P^i}} + \delta_{P^i, t^*-1}
\end{equation}
Although written as a single term, the second term in \Cref{equ:downstream-components} may be separated into multiple updates in both the prefetch scheduling and actual prefetch process to take full advantage of the prefetch budget.

The first term, $\overline{w_{P^i}}$, is unavoidable. So, the additional $\delta_{P^i, t^*-1}$ term is the main culprit for bandwidth overhead. 
Moreover, since the client model at the start of the \emph{Train} phase using \myname is equivalent to the client model without prefetch, the convergence behavior of \myname should be similar to the non-prefetch case. Therefore, the benefits brought by \myname will increase with stronger downstream compressors $C_{dl}$. 

\begin{algorithm}[t!]
\caption{\myname}\label{alg:FedFetch}
\begin{algorithmic}[1]
\Procedure{Server}{}
    \For{$t \gets 1, \dots, T$}
        \LineCommentBlue{Server: \emph{Prepare} phase}
        \State Presample $\mathcal{K}_{t+R}$ from $\mathcal{N}$
        \State $P_{t+R} \gets \textsc{SchedulePrefetch}()$
        \LineCommentBlue{Clients in $\mathcal{K}_{t+R}$ start \emph{Prefetch} phase}
        \LineCommentBlue{Clients in $\mathcal{K}_t$ start \emph{Train} phase}
        \LineCommentBlue{Server: Aggregation}
        \State $\Delta_t \gets \sum_{i\in \mathcal{K}_{t}}\nu_{i}\hat{\Delta}_{t}^i$ \label{$alg:FedFetch:ServerAggregation$}
        \State $w_{t+1} \gets w_t + C_{dl}(\Delta_t)$
    \EndFor
\EndProcedure
\State
\Procedure{Client $i$ }{}
    \ParState{\textcolor{blue}{\(\triangleright\)  Server notification happens on scheduled round $P^i$. Clients receive the current round $t$ 
    and train round $t^*$}}
    \State \textsc{NotifiedByServer}()
    \LineCommentBlue{Client: \emph{Prefetch} phase ($t<t^*$)}
    \LineCommentBlue{Sync base model}
    \State $\hat{w}_{\ell_i}^i \gets \textsc{Download}(\overline{w_{P^i}})$ 
    \label{alg:sync-base-model}
    \State $\ell_i \gets P^i - 1$
    
    \LineCommentBlue{Sync server updates}
    \While{$t \gets \textsc{QueryServerRound}()$, $t < t^*$}
        \State $\hat{w}_t^i \gets \hat{w}_{\ell_i}^i + \textsc{Download}(\delta_{\ell_i, t-1})$ \label{alg:sync-server-updates}
        \State $\ell_i \gets t-1$ 
    \EndWhile 
    \LineCommentBlue{Client: \emph{Train} phase ($t=t^*$)}
    \State $\hat{w}_t^i \gets \hat{w}_{\ell_i}^i + \textsc{Download}(\delta_{\ell_i, t-1})$ \label{alg:FedFetch:train-phase-updates}
    \State $\hat{\Delta}_{t}^i \gets C_{dl}(\textsc{LocalTraining}(\hat{w}_t^i) - \hat{w}_t^i)$
    \State $\textsc{Upload} (\hat{\Delta}_{t}^i)$
\EndProcedure
\end{algorithmic}
\end{algorithm}

\subsection{Prefetch Phase}
\label{sec:prefetch}
Once client $i$ from the presampled client set $K_{t^*}$ acquires its prefetch schedule $P^i$ from the \emph{Prepare} phase, the client can start its \emph{Prefetch} phase. Client $c_i$ will attempt to prefetch starting at round $P^i$ until the \emph{Train} phase at round $t^*$. 

\subsubsection{Prefetch Process}
\myname employs a greedy prefetch process where the presampled client $c$ always tries to download the most recently available update starting on their scheduled prefetch start round $P^c$.
To minimize the \emph{Train} phase fetch size and time, \myname transmits the largest downstream updates at the start and the smallest updates ($\delta$ of a single round) at the end of the \emph{Prefetch} phase.

\label{sec:prefetch-process}
\begin{figure}
    \centering
    \includegraphics[width=1.0\linewidth]{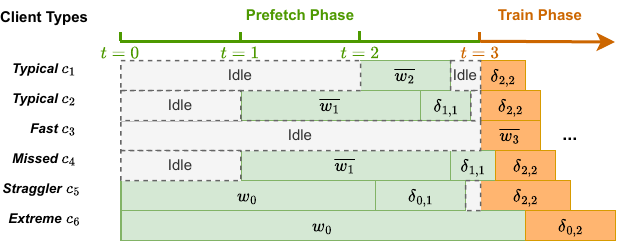}
    \caption{An example of a prefetch process for six clients and $R=3$. The blocks represent what each client is currently prefetching (in green) or fetching (in orange) from the server.
    } 
    \label{fig:prefetch-process}
\end{figure}

\Cref{fig:prefetch-process} illustrates an example \myname run with 6 clients of different categories and $R=3$. On round 0, \myname presamples all 6 clients who will start training on round 3. \myname also provides a prefetch schedule for each client which is indicated by the round a client $c_i$ starts prefetching.

A typical client like $c_1$ will start prefetching on round $2$ and completes before the \emph{Train} phase in round $3$. TRhe update acquired by the client in its \emph{Train} phase is the smallest possible update $\delta_{2,2}$.
In contrast, another client, $c_2$, needs two rounds to prefetch all its updates. Therefore, $c_2$ prefetches in round $1$
and continues to prefetch $\delta_{1,1}$ as its local model is from the start of round 1 and the new update $\delta_{1,1}$ is now available. 

For a client with substantial bandwidth, $c_3$, the server assigns it to start prefetching on round $3$ because $c_3$ can finish downloading the full model $\overline{w_3}$ in its \emph{Train} phase. 
However, clients like $c_4$ may not prefetch all their \emph{Prefetch} phase updates before round 3. This forces client $c_4$ to acquire both pending prefetch and \emph{Train} phase updates, increasing fetch time. 

Stragglers like $c_5$ require the maximum prefetch budget of $R=3$ rounds to prefetch its updates starting from round $0$. Clients with low bandwidth, like $c_6$, may still fail to finish prefetching within 3 rounds. But, prefetching benefits such clients by reducing the volume they download during the \emph{Train} phase compared to cases with no prefetching.

\subsubsection{Impact of Client Unavailability}

In real deployments of cross-device FL, clients are online at different times. Therefore, client sampling process only samples from the set of clients that are currently online. 
So, a method that selects clients earlier than normal may suffer performance drops due to clients going offline. \myname addresses this problem with a simple replacement strategy. These replacement clients will immediately start prefetching if there is still a prefetch budget. We observe empirically that this mostly negates the performance issues caused by unavailability (\Cref{sec:availability-analysis}). 

\section{Experimental Evaluation}
\begin{table*}[t!]

\caption[Caption for Performance Results]{Main performance results for three model-dataset pairs. Metrics \textbf{FT} and \textbf{TT} represent \textbf{F}etch \textbf{T}ime and \textbf{T}otal Training \textbf{T}ime, in hours. \textbf{FV} and \textbf{TV} represent \textbf{F}etch \textbf{V}olume and \textbf{T}otal Transmission \textbf{V}olume, in $\times10^2$ GB. 
Results are recorded when the target accuracy (\textit{Trg}) is reached\textsuperscript{a}. We run each setting at least 3 times and report the mean.}
\small\textsuperscript{a} PowerSGD achieves a maximum test accuracy of 61\% on the OpenImage dataset, we \underline{underline} these results

\label{tab:perf-results}
\begin{center}
\begin{tabular}{r|r|llll|llll|llll}
\specialrule{.1em}{.05em}{.05em} 
\multicolumn{2}{r|}{
}
& \multicolumn{4}{c|}{\emph{FEMNIST Trg 75\%}} & \multicolumn{4}{c|}{\emph{Google Speech Trg 61\%}} & \multicolumn{4}{c}{\emph{OpenImage Trg 68\%}} \\ 
\multicolumn{2}{r|}{
}                          & \textbf{FT}   & \textbf{TT}  & \textbf{FV}  & \textbf{TV}  & \textbf{FT}    & \textbf{TT}   & \textbf{FV}   & \textbf{TV}  & \textbf{FT}    & \textbf{TT}   & \textbf{FV}   & \textbf{TV} \\ 
\specialrule{.1em}{.05em}{.05em} 
\textbf{Baseline}                      & FedAvg       & 0.64 & 2.56 & 1.56 & 2.76 & 5.54  & 21.7 & 12.2 & 21.6 &1.14  &5.4  &10.8 & 19.1\\ \hline
\multirow{4}{*}{\textbf{Masking}}      & STC          & 0.85 & 2.46 & 2.58&\textbf{3.07} & 9.96   & 25.0  & 13.7 & \textbf{18.0} & 0.97  & 4.47  & 11.5 & \textbf{15.0}\\
                              & \myname + STC 
                                                      & \textbf{0.21} & \textbf{1.67} & \textbf{1.31} & 3.14 & \textbf{2.71}   & \textbf{13.4} & \textbf{10.4} & 21.6 & \textbf{0.44}  & \textbf{4.00} &\textbf{8.85} & 18.1 \\
                               \cline{2-14}
                              & GlueFL                & 0.30 & 2.11 & 2.58&\textbf{3.32} & 2.28 & 16.7 & 12.9 & \textbf{18.9} & 0.38 & 2.55 & 14.0 & \textbf{20.7} \\
                              & \myname + GlueFL      & \textbf{0.18} & \textbf{1.84} & \textbf{1.49} & 4.01 & \textbf{1.26} & \textbf{14.3} & \textbf{11.2}  & 25.0 & \textbf{0.23} & \textbf{2.36}  & \textbf{8.84} & 24.6\\ \hline
\multirow{6}{*}{\textbf{Quantization}} & QSGD         & 0.46 & 1.35 & 1.56 &\textbf{1.73}& 3.41   & 8.62 & 10.49 & \textbf{11.6} &0.72  &3.68 &10.8&\textbf{12.0}\\
                              & \myname + QSGD        & \textbf{0.06} &\textbf{0.90} & \textbf{0.58}& 1.83& \textbf{0.84}  & \textbf{6.26}  &  \textbf{4.87} & 14.3 &\textbf{0.13}&\textbf{3.07} &\textbf{4.45} &13.7\\
                              \cline{2-14}
                              & LFL                   & 0.43 & 1.27 & 1.45 & \textbf{1.60}& 3.03   & 10.9 & 10.78 & \textbf{11.9} &0.67&3.55 &10.4&\textbf{11.6}\\
                              & \myname + LFL         & \textbf{0.06}& \textbf{0.90} & \textbf{0.58} & 1.83& \textbf{0.58}   & \textbf{8.88}  &  \textbf{4.7} & 14.7 &\textbf{0.12}&\textbf{3.09}&\textbf{4.34}&13.7\\
                              \cline{2-14}
                              & EDEN                  & 0.42 & 1.26 &1.43 & \textbf{1.60} & 4.02   & 12.7 & 11.7 &  \textbf{13.7}  &0.74&3.54&10.2&\textbf{11.7}\\
                              & \myname + EDEN        & \textbf{0.06} & \textbf{0.89} & \textbf{0.60}& 1.75 & \textbf{1.12}  & \textbf{10.1} & \textbf{6.92}  &  16.5  &\textbf{0.16}&\textbf{3.12}&\textbf{5.48}&14.1\\ \hline

\multirow{2}{*}{\textbf{Low-rank}}     & POWERSGD      & 1.49 & 4.23 & 5.25 & \textbf{5.62}& 6.87 & 26.7  &  28.1  & \textbf{29.8}  & \underline{0.58} & \underline{2.81}  & \underline{8.01}  & \underline{\textbf{9.24}} \\
                              & \myname + POWERSGD     & \textbf{0.14} & \textbf{3.02}& \textbf{1.47}& 6.54&  \textbf{0.67}& \textbf{21.0} & \textbf{6.89} & 35.5 & \underline{\textbf{0.11}}  & \underline{\textbf{2.43 }} & \underline{\textbf{4.28}} & \underline{11.0} \\ \hline
\textbf{Quantization}                  & DoCoFL                & 0.04 & 0.96& \textbf{0.13} &0.72 &  \textbf{0.48 } & 8.79 & \textbf{1.45}  & 6.97 &0.12&3.26&\textbf{1.15}&5.52\\
\textbf{(Full model)}                  & \myname + DoCoFL       & 0.04 & \textbf{0.93}& 0.28 &\textbf{0.56} &  0.50  & \textbf{8.47} & 2.95  &\textbf{6.02}  &0.12&\textbf{2.87}&2.50&\textbf{5.38}\\
\specialrule{.1em}{.05em}{.05em} 
\end{tabular}
\end{center}
\end{table*}

We evaluate \myname along several dimensions and answer the following four questions:

\begin{itemize}[\IEEEsetlabelwidth{Z}]
    \item[Q1:] How well does \myname integrate with existing techniques such as client sampling, masking, quantization, and low-rank  decomposition?
    \item[Q2:] What impact does \myname have on training time, bandwidth usage, and model accuracy?
    \item[Q3:] How does FedFetch’s hyperparameter impact its performance?
    \item[Q4:] How does \myname handle settings involving overcommitment and client unavailability?
\end{itemize}

\subsection{Experimental Setup}
\label{sec:experiment-setup}
\subsubsection{Environment and Datasets}
We run all experiments on the FedScale\cite{fedscale-icml22} platform.
We use client bandwidth data points from the Measurement Lab's NDT data set~\cite{mlab} (\Cref{fig:speed-test}). 
We rely on FedScale to organize client device hardware data from AI Benchmark~\cite{aibench} and online/offline behaviour traces from FLASH~\cite{yang2021characterizingHeterogeneity} (\Cref{sec:availability-analysis}).

We use benchmarking datasets provided by FedScale: FEMNIST~\cite{caldas2019leaf}, Google Speech~\cite{warden2018speech}, and OpenImage~\cite{OpenImages}. The FEMNIST and OpenImage datasets are used to train image classification models. The former consists of 640K colored images and 2,800 clients and the latter consists of 1.3M colored images and 10,625 clients. The Google Speech dataset is used to train a speech recognition model and consists of 105K speech samples and 2,066 clients. We train different models --- FEMNIST uses ShuffleNet~\cite{zhang2018shufflenet}, Google Speech uses ResNet-34~\cite{he2016deep}, and OpenImage uses MobileNet~\cite{sandler2018mobilenetv2}. Similar to recent work~\cite{fedscale-icml22}, we set the target accuracy to be the highest achievable accuracy by tested methods. The number of client results collected per round is $K = 30$ for FEMNIST and Google Speech, and $K = 100$ for OpenImage. We set over-commitment $OC=1.3$~\cite{bonawitz2019FLAtScale}; the actual number of clients sampled per round will increase by $1.3 \times$.

\subsubsection{Compression strategies}
\myname works with most existing compression methods applied to FedAvg with simple random client sampling~\cite{mcmahan2017FedAvg}. To answer Q1, we compare the method with FedAvg versus using \myname and the method together. We use STC\footnote{STC is a hybrid method featuring both masking and quantization which are orthogonal techniques, we exclusively evaluate STC's masking strategy}~\cite{sattler2019STC}, GlueFL\footnote{We replace simple random sampling with the sticky client sampling}
~\cite{he2023gluefl} as the representatives for masking, QSGD~\cite{alistarh2017qsgd}, LFL~\cite{LFL-amiri2020}, EDEN~\cite{EDEN-pmlr-v162-vargaftik22a} for quantization, and PowerSGD~\cite{vogels2019powersgd} for matrix decomposition. We also evaluate \myname's compatibility with DoCoFL~\cite{DoCoFL-pmlr-v202-dorfman23a}, a full model quantization method 
by replacing DoCoFL's transmission of models from the top of a compressed model queue with \myname's prefetch method, which sends more recent compressed server models.

\subsubsection{Parameters} 
\label{sec:parameters}
We use parameter values from previous works which reliably reach the target test accuracy. Clients perform 10 local updates per round with PyTorch's SGD optimizer with a momentum factor of 0.9 on batches of size 20. The initial learning rate is set to 0.01
and decreases by 0.98 every 10 rounds. For STC and GlueFL, we set a compression ratio of $q = 20\%$ for ShuffleNet, $q=35\%$ for MobileNet, and $q=30\%$ for ResNet-34. 
For quantization, we use a bit budget of 4 bits for QSGD, LFL, and EDEN. For PowerSGD, we set the rank to $16$ for ShuffleNet and $24$ for MobileNet and ResNet-34. Finally, we set DoCoFL's bit-budget, anchor deployment rate, and anchor queue size to 2, 10 and 3 respectively.

\subsubsection{Metrics}
For Q2 and Q3, we measure the end-to-end time, download-specific training time, and bandwidth usage.
Since the straggler client will determine the total time duration of an FL round, we use the sum of the stragglers' download times to represent the total fetch time (FT) and similarly for compute and upload time. This way, the total FL training time (TT) is equal to the sum of the fetch (FT), compute, and upload times. For bandwidth usage, the total transmission volume (TV) represents the transmission volume used by every client, including overcommitted clients. The fetch volume (FV) represents the downstream bandwidth associated with the fetch operation for all clients participating in the \emph{Train} phase.
For Q4, we report the time and bandwidth  when the average of the previous 5 testing rounds reaches the target accuracy listed in~\Cref{tab:perf-results}.
This is similar to prior work~\cite{oort-osdi21, he2023gluefl}. 

\subsection{Results }

\subsubsection{Main Performance Results (Q1--Q2)}
\label{sec:main-perf}

Using settings from the previous section, we experiment with each compression method with and without \myname. Therefore, \myname's baselines are the compression methods without \myname. After reaching the target accuracy in \Cref{tab:perf-results}, we record the time-to-accuracy and bandwidth-to-accuracy performance.

\Cref{tab:perf-results} shows that \myname consistently saves fetch (download) and end-to-end training time at little extra bandwidth overhead for all tested compression algorithms and across different models and datasets. On average, we see a 4.49$\times$ reduction in fetch time. This translates into a mean end-to-end training time speedup of 1.26$\times$. 

The time savings are significant for masking techniques like STC, where \myname speeds up training time by 1.49$\times$. \myname is less effective at improving GlueFL (1.09$\times$ end-to-end speedup on average). However, this is expected because GlueFL already employs sticky sampling, a client sampling technique that favors clients who recently participated in training. For quantization methods like QSGD, LFL, and EDEN, \myname achieves an average end-to-end time speedup of 1.29$\times$. We further show that \myname reduces PowerSGD's end-to-end time by 1.28$\times$. The total time savings (1.07$\times$ speedup) are less noticeable for full model quantization methods like DoCoFL. 
This is expected because DoCoFL already leverages a simple form of prefetching which we optimize further with \myname. Specifically, through prefetch scheduling, \myname allows DoCoFL to safely use more recently compressed base models without impacting round duration whereas DoCoFL uses a more stale model from the top of its compressed model queue. 

Crucially, \myname achieves the above speedups with a 13\% average increase in bandwidth. \myname fulfills its primary goal of speeding up downstream compression communication in cross-device FL for a broad range of compression techniques in a bandwidth-efficient manner.

\subsubsection{Bandwidth Breakdown (Q2)}
\label{sec:bw-usage}
\begin{figure}
    \centering
    \includegraphics[width=\linewidth]{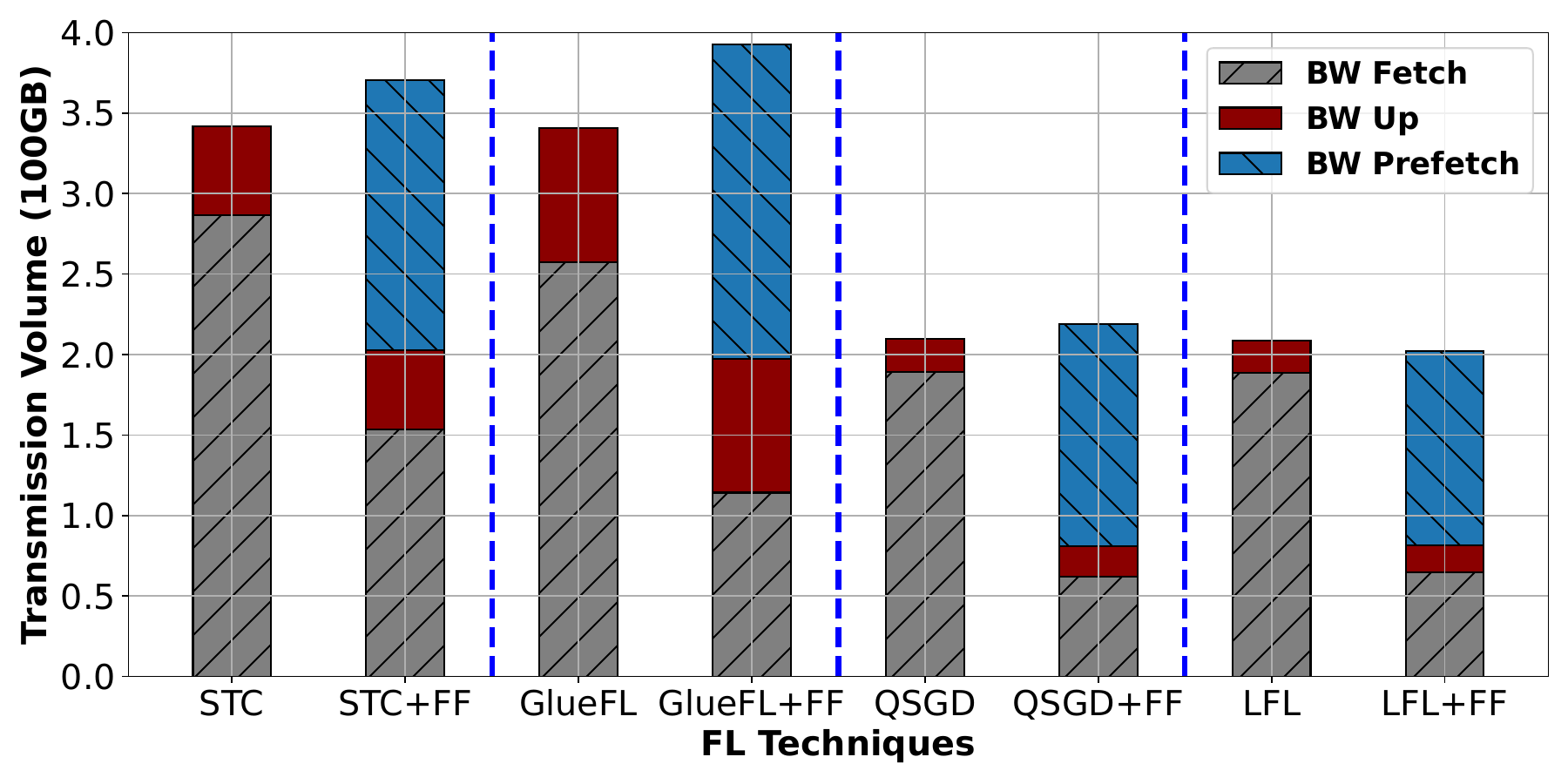}
    \caption{Bandwidth usage of select FL techniques with and without \myname. Techniques ending with ``+FF" apply \myname. Each bar is divided into Fetch, Up(load), and Prefetch.}
    \label{fig:bw-usage}
\end{figure}

We examine the effect of \myname on bandwidth usage in \Cref{fig:bw-usage}, which plots the total volume breakdown between prefetch, fetch, and upstream for STC, GlueFL, LFL, QSGD methods with and without \myname after they reach the target accuracy for FEMNIST. 
We see a shift in the distribution from fetch to prefetch when \myname is used. With \myname, the fetch bandwidth is, on average, 59\% lower. This suggests that \myname achieves the goal of shifting downstream bandwidth from fetch to prefetch.

\subsubsection{Comparison with Naive Forms of Prefetching (Q2)}
\label{sec:naive-prefetch-analysis}
\begin{figure}[t]
     \centering
     \begin{subfigure}[b]{0.49\linewidth}
         \centering
         \includegraphics[width=\linewidth]{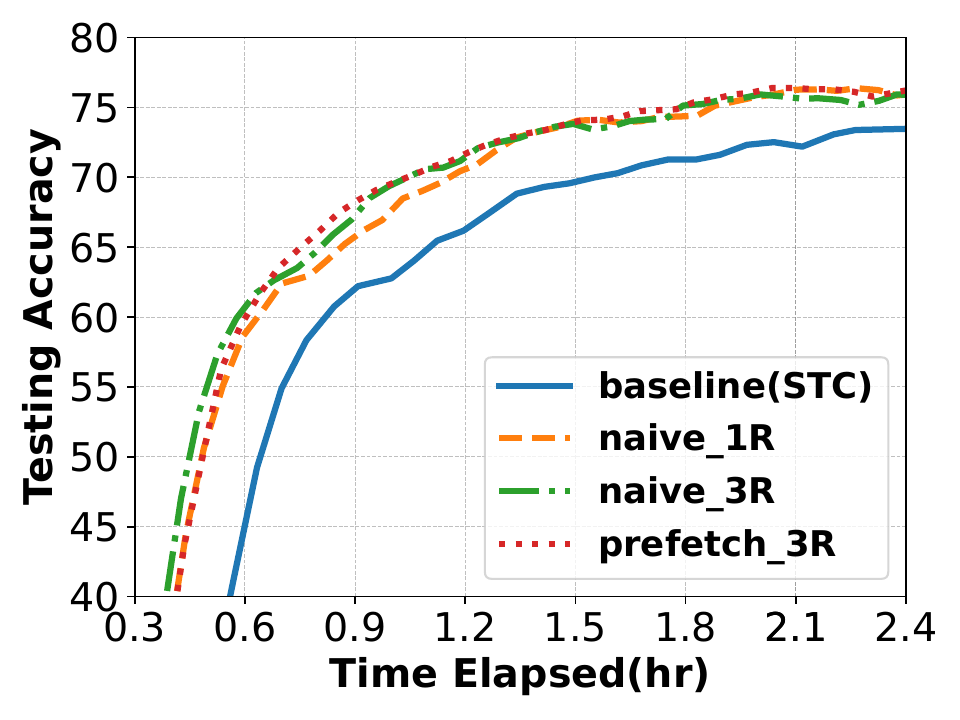}
         \caption{Comparison with naive prefetching on time-to-accuracy performance for STC.}
         \label{fig:naive-time-STC}
     \end{subfigure}
     \hfill
     \begin{subfigure}[b]{0.49\linewidth}
         \centering
         \includegraphics[width=\linewidth]{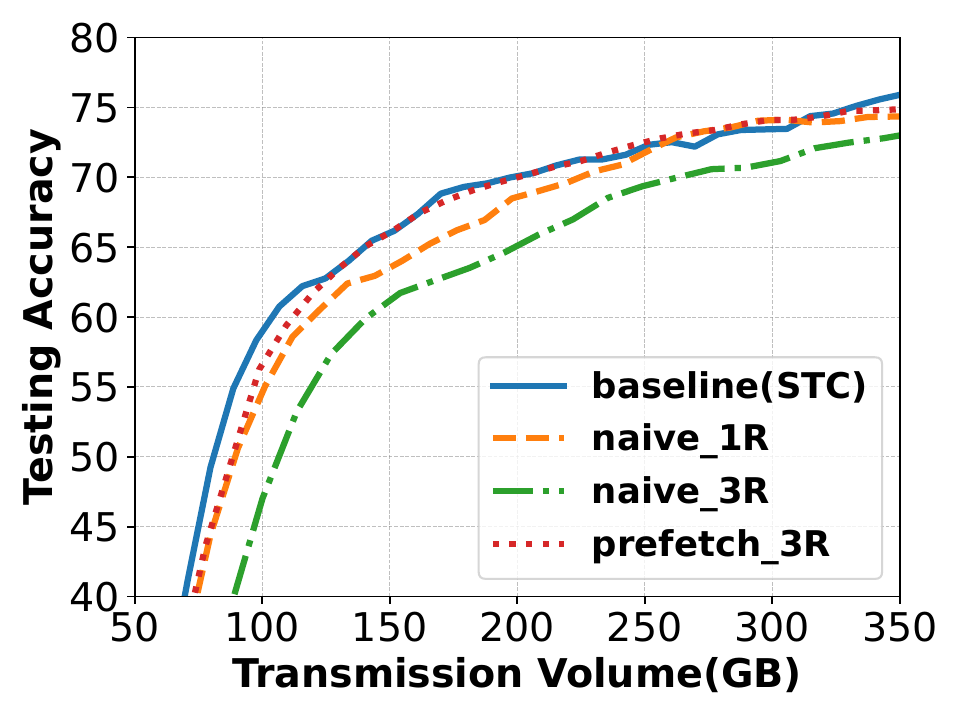}
         \caption{Comparison with naive prefetching on bandwidth-to-accuracy performance for STC.}
         \label{fig:naive-bandwidth-STC}
     \end{subfigure}
     \begin{subfigure}[b]{0.49\linewidth}
         \centering
         \includegraphics[width=\linewidth]{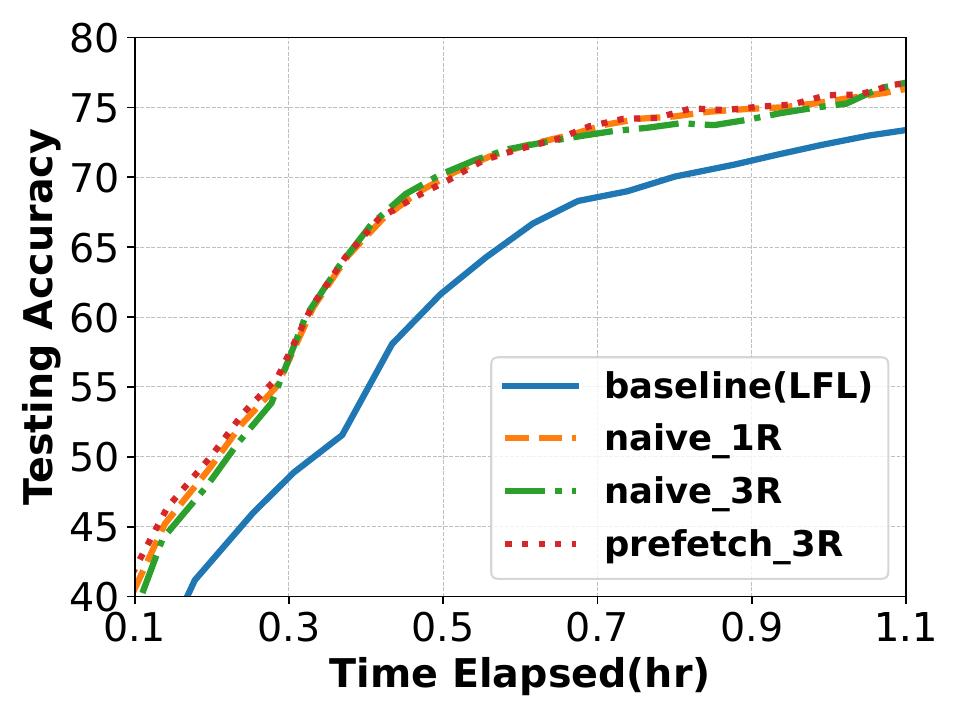}
         \caption{Comparison with naive prefetching on time-to-accuracy performance for LFL.}
         \label{fig:naive-time-LFL}
     \end{subfigure}
     \hfill
     \begin{subfigure}[b]{0.49\linewidth}
         \centering
         \includegraphics[width=\linewidth]{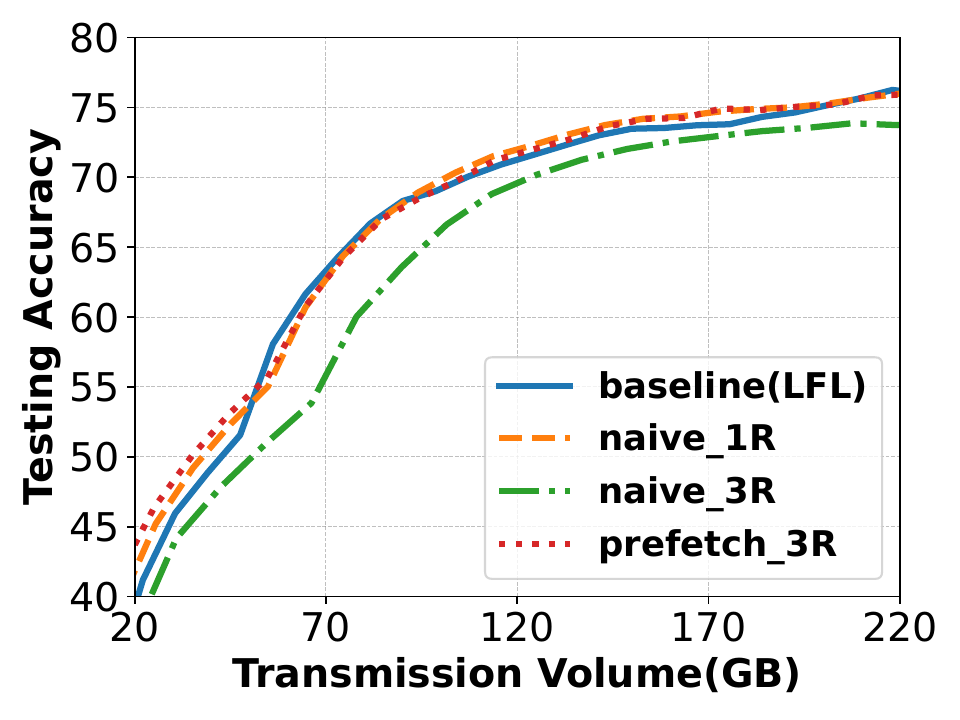}
         \caption{Comparison with naive prefetching on bandwidth-to-accuracy performance for LFL.}
         \label{fig:naive-bandwidth-LFL}
     \end{subfigure}
    \caption{Comparison with naive forms of prefetching for masking with STC and quantization with LFL.}
    \label{fig:naive-analysis}
\end{figure}
We compare \myname with two simple prefetching methods on FEMNIST. In the first, every presampled client shares a fixed prefetch schedule of 1 round, while the second method uses 3 rounds. We plot the time and total bandwidth versus accuracy performance for the base compression method (STC or LFL), \myname with $R=3$, and the two simple prefetching methods in \Cref{fig:naive-analysis}.  These plots show that \myname brings as much time-to-accuracy performance as using a fixed 3-round prefetch schedule. But, \myname uses the same or less amount of extra bandwidth as the fixed 1-round prefetching, despite a prefetch budget of 3 rounds. \myname achieves the maximum speedup for a given $R$ without extra bandwidth usage.

\subsubsection{Sensitivity Analysis of $R$ (Q3)}
\label{sec:sensitivity-analysis}
\begin{figure}
     \centering
     \begin{subfigure}[b]{0.49\linewidth}
         \centering
         \includegraphics[width=\linewidth]{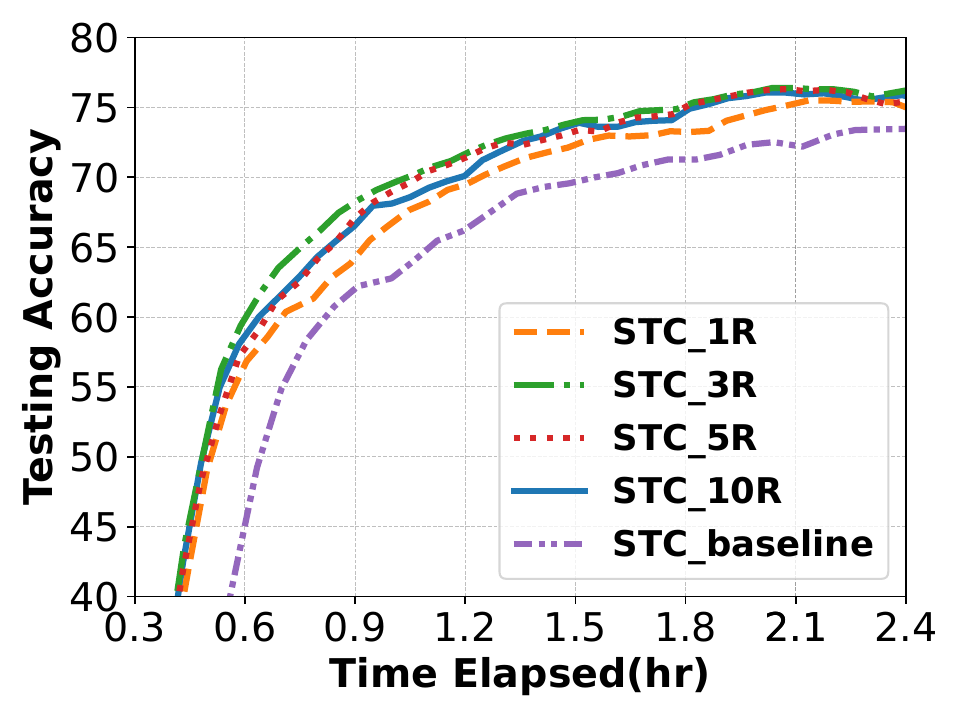}
         \caption{Effect of $R$ on time-to-accuracy performance for STC.}
         \label{fig:sensitivity-time-stc}
     \end{subfigure}
     \hfill
     \begin{subfigure}[b]{0.49\linewidth}
         \centering
         \includegraphics[width=\linewidth]{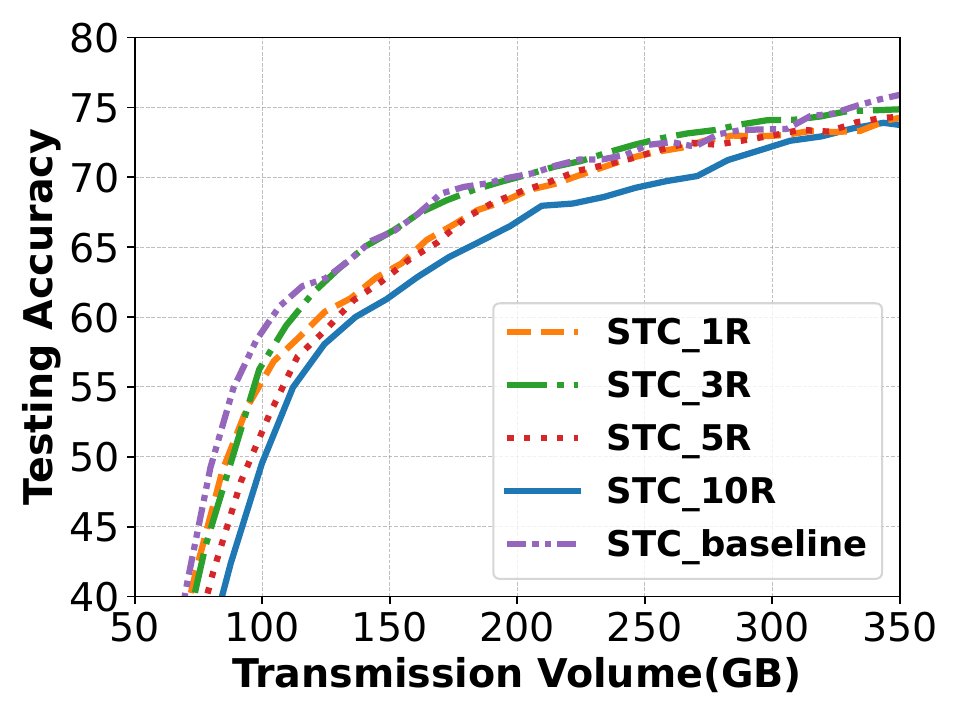}
         \caption{Effect of $R$ on bandwidth-to-accuracy performance  for STC.}
         \label{fig:sensitivity-bandwidth-stc}
     \end{subfigure}
     \begin{subfigure}[b]{0.49\linewidth}
         \centering
         \includegraphics[width=\linewidth]{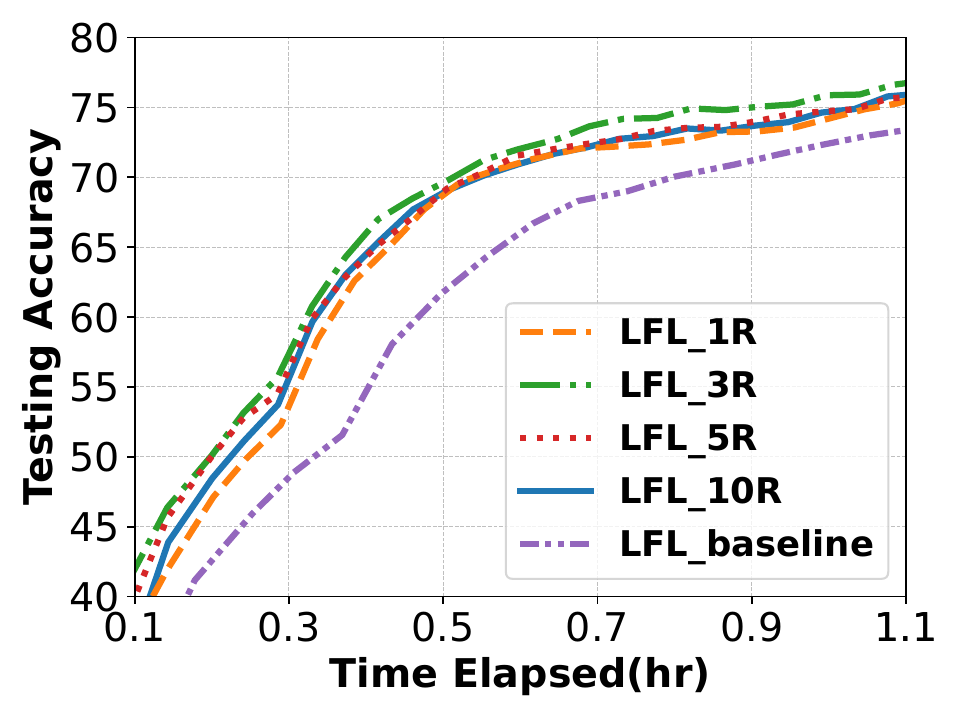}
         \caption{Effect of $R$ on time-to-accuracy performance  for LFL.}
         \label{fig:sensitivity-time-lfl}
     \end{subfigure}
     \hfill
     \begin{subfigure}[b]{0.49\linewidth}
         \centering
         \includegraphics[width=\linewidth]{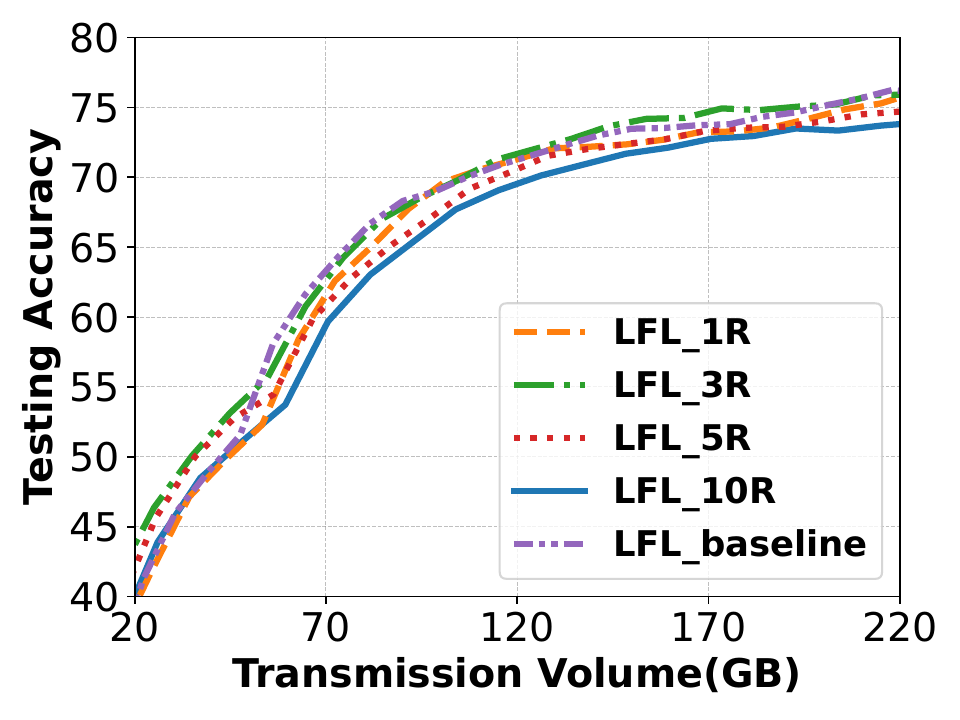}
         \caption{Effect of $R$ on bandwidth-to-accuracy performance for LFL.}
         \label{fig:sensitivity-bandwidth-lfl}
     \end{subfigure}
     
    \caption{Sensitivity analysis for the max number of prefetch rounds $R$ for masking with STC and quantization with LFL.}
    \label{fig:sensitivity}
\end{figure}

We evaluate the impact of the max number of prefetch rounds, $R$, on performance. We apply \myname to STC and LFL on FEMNIST. We vary $R$, choosing values of 1, 3, 5, and 10.
For any choice of $R$, adding \myname to a compression method shifts the corresponding time-to-accuracy curve to the left, as seen in \Cref{fig:sensitivity-time-stc}. This indicates that adding \myname can consistently reduce training time. Among these, the curve associated with $R=1$ has the worst time-to-accuracy performance. This is expected because the slowest selected clients may require more than 1 round to complete prefetching. For $R > 1$, the difference in time-to-accuracy and bandwidth-to-accuracy performance is smaller because stragglers needing multiple rounds of prefetching participate less frequently in training. Nevertheless, \Cref{fig:sensitivity-bandwidth-stc} and \Cref{fig:sensitivity-bandwidth-lfl} show that the bandwidth consumption of \myname only increases slightly for larger values of $R$. This indicates that \myname can minimize the extra bandwidth overhead associated with prefetching.

\subsubsection{Sensitivity Analysis of $\beta$ (Q3)}
\label{sec:beta-analysis}
\begin{figure}
    \centering
    \centering
     \begin{subfigure}[t]{0.49\linewidth}
         \centering
         \includegraphics[width=\linewidth]{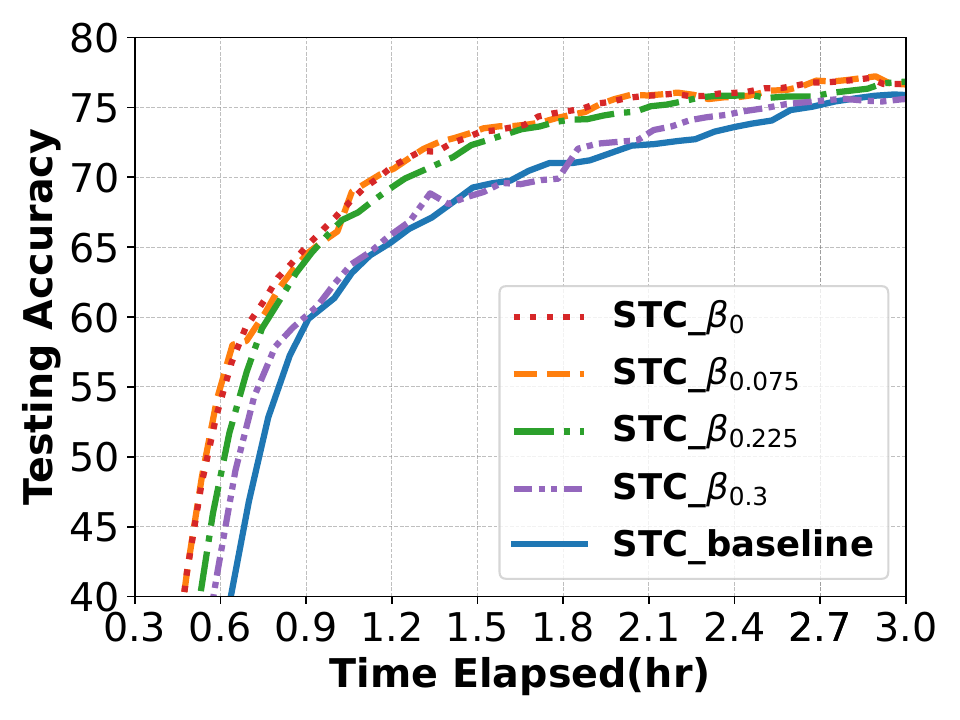}
         \caption{Impact of $\beta$ on time-to-accuracy performance.}
         \label{fig:beta-analysis-time}
     \end{subfigure}
     \hfill
     \begin{subfigure}[t]{0.49\linewidth}
         \centering
         \includegraphics[width=\linewidth]{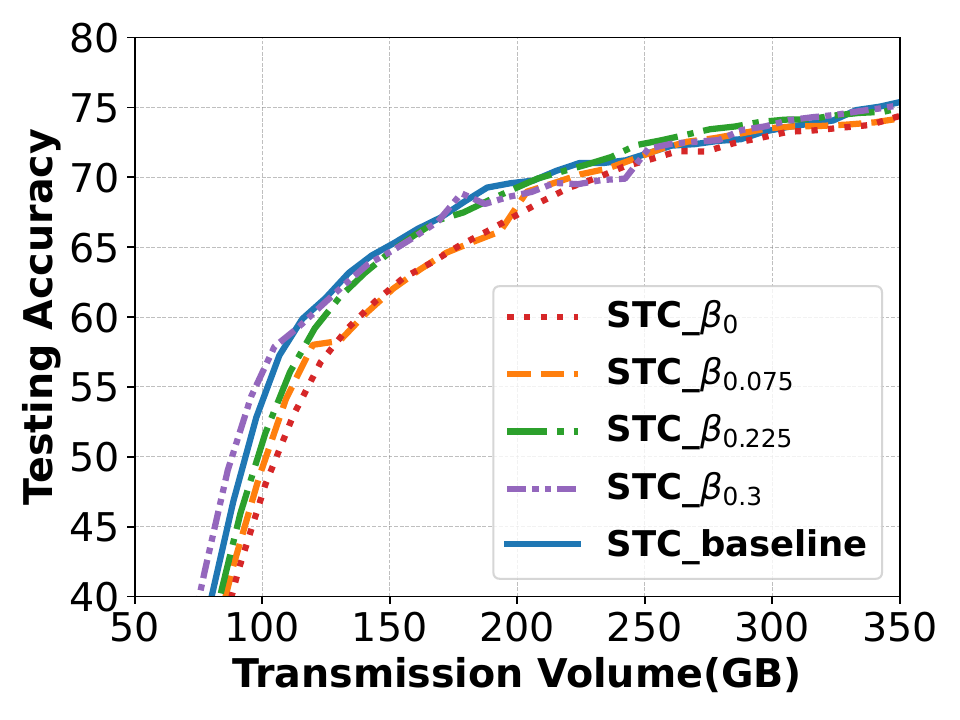}
         \caption{Impact of $\beta$ on bandwidth-to-accuracy performance.}
         \label{fig:beta-analysis-bw}
     \end{subfigure}
    \caption{Impact of $\beta$ for \myname + STC.}
    \label{fig:beta-analysis}
\end{figure}

The prefetch scheduler's factor $\beta$ (see \Cref{sec:preample-schedule-estimation}) controls how aggressive the prefetch scheduler is with scheduling clients with less rounds of prefetching. Under the standard overcommitment $OC=1.3$ setting, we plotted the time and bandwidth to accuracy of  $\beta$ choices in \Cref{fig:beta-analysis} as well as the baseline STC setting without \myname. A higher $\beta$ leads to more greater proportion of clients prefetching at an earlier round. The similar performance of $\beta =0, 0.075, 0.225$ in \Cref{fig:beta-analysis-time} suggests that $\beta$'s effect on \myname's time improvements is limited, unless $1+\beta$ is close to $OC$. Meanwhile, \Cref{fig:beta-analysis-bw} demonstrates that lower values of $\beta$ do incur additional transmission volume at lower testing accuracies but the effect becomes insignificant at higher accuracies. One interesting case occurs when $\beta=0.3$ so $1+\beta = 1.3 = OC$ where only the slowest client perform prefetching. However, since this client is highly likely to remain a straggler even after prefetching, the over-commitment mechanism may discard this client. In effect, $\beta = 0.3$ resembles the no-prefetch baseline in environments with real-life heterogeneous bandwidth distributions.

\subsubsection{
Sensitivity Analysis of $\alpha$ (Q3)}
\label{sec:alpha-analysis}
\begin{figure}
    \centering
    \centering
     \begin{subfigure}[t]{0.49\linewidth}
         \centering
         \includegraphics[width=\linewidth]{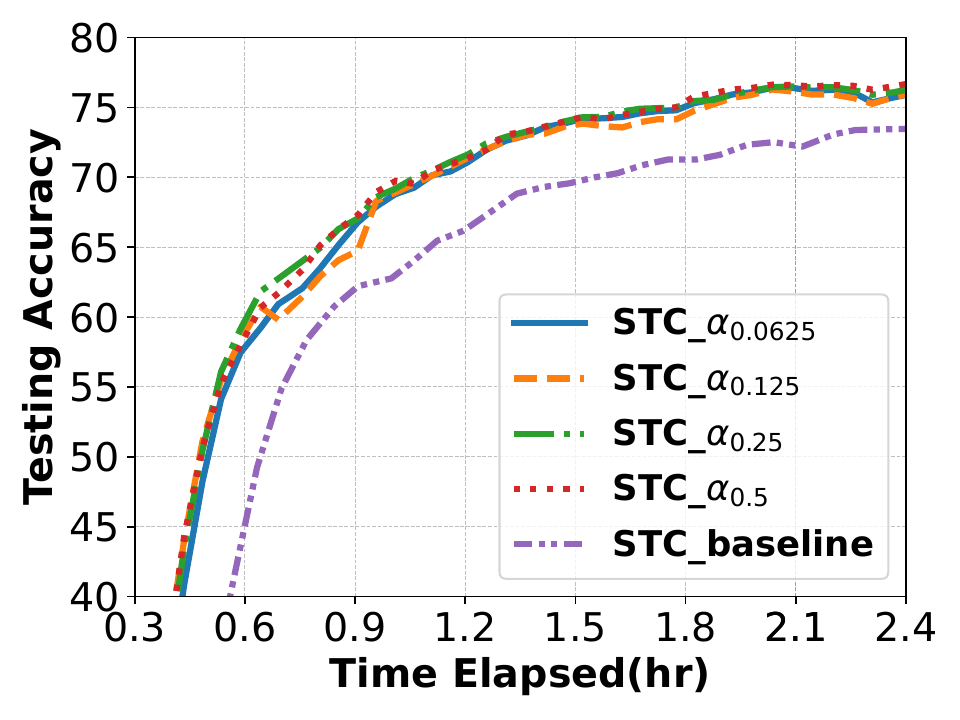}
         \caption{Impact of $\alpha$ on time-to-accuracy performance.}
         \label{fig:alpha-analysis-time}
     \end{subfigure}
     \hfill
     \begin{subfigure}[t]{0.49\linewidth}
         \centering
         \includegraphics[width=\linewidth]{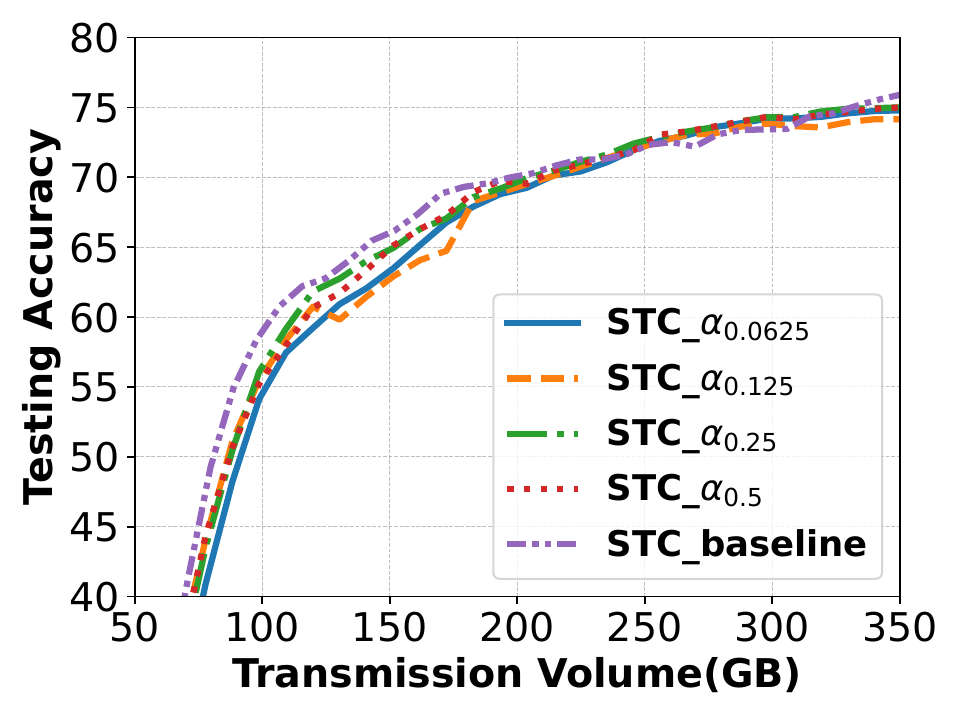}
         \caption{Impact of $\alpha$ on bandwidth-to-accuracy performance.}
         \label{fig:alpha-analysis-bw}
     \end{subfigure}
    \caption{Impact of $\alpha$ for \myname + STC.}
    \label{fig:alpha-analysis}
\end{figure}
The smoothing factor $\alpha$ represents the relative importance of the more recently observed round duration $d_t$ when calculating the moving average in \Cref{equ:duration-estimate}. In \Cref{fig:alpha-analysis}, we evaluate various settings of $\alpha$, including the default value of $\alpha=0.125$. FedFetch's performance across $\alpha$ values do not differ significantly. Due to the over-commitment setting of $OC=1.3$, the FL training already removes extreme stragglers which limits the amount of variance in round durations.

\begin{table}[t]
\caption{Impact of overcommitment $OC$ on the FEMNIST dataset (Trg 75\%). See \Cref{tab:perf-results} for column definitions.}
\label{tab:OC}
\centering
\begin{tabular}{r|cccc|cccc}
\specialrule{.1em}{.05em}{.05em} 
     & \multicolumn{4}{c|}{STC} & \multicolumn{4}{c}{STC + FedFetch} \\
    $OC$&\textbf{FT}&\textbf{TT}&\textbf{FV}&\textbf{TV}&\textbf{FT}&\textbf{TT}&\textbf{FV}&\textbf{TV}\\
     \hline
     1.0 & 10.97& 51.33& 1.97& \textbf{2.45} & \textbf{5.5} & \textbf{48.39} & \textbf{1.90}&2.63\\
     1.1 &  3.03&  8.12& 1.77& \textbf{2.37} & \textbf{1.16}& \textbf{7.34} & \textbf{1.50}&2.68\\
     1.2 &  1.54&   3.5& 2.09& \textbf{2.53} & \textbf{0.59}& \textbf{2.75} & \textbf{1.44}&3.09\\
     1.3 &  0.85&  2.46& 2.58& \textbf{3.07} & \textbf{0.21}& \textbf{1.67} & \textbf{1.31}&3.14\\
     1.4 &  0.57&   2.04& 2.78& 3.28 & \textbf{0.11}& \textbf{1.37} & \textbf{1.23}&\textbf{3.22}\\
\specialrule{.1em}{.05em}{.05em} 
\end{tabular}
\end{table}

\subsubsection{
Impact of Over-commitment (Q4)}
In~\Cref{tab:OC}, as over-commitment $OC$ increases, more clients with weaker connectivity are removed. This leads to faster training time at the cost of higher data transmission volume, with this effect the most significant for lower values of $OC$.
\myname consistently decreases FT, TT, and FV across $OC$ values. Note that the fetch volume decreases with $OC$ values for STC+FedFetch instead of increasing as in STC. This is because clients with average bandwidth are more numerous and are likely to become stragglers under higher $OC$ settings. \myname's adaptively encourages prefetch schedules with more prefetch rounds for these clients. This shifts bandwidth consumption from fetch to prefetch.

\subsubsection{Impact of Client Availability (Q4)}
\label{sec:availability-analysis}
\begin{figure}
    \centering
    \centering
     \begin{subfigure}[t]{0.49\linewidth}
         \centering
         \includegraphics[width=\linewidth]{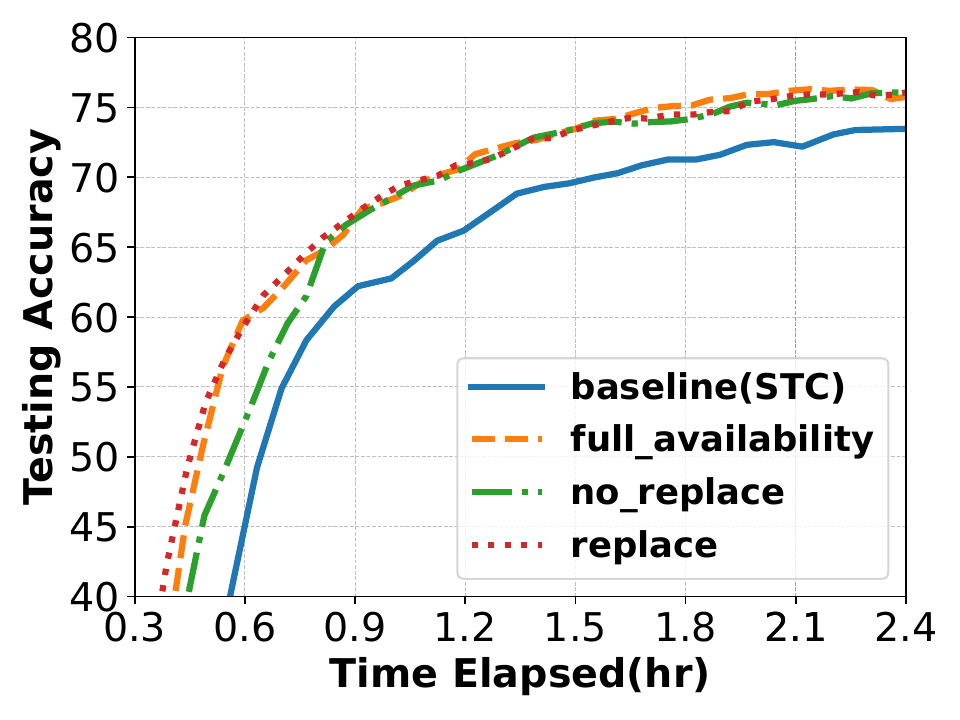}
         \caption{Impact of client availability on time-to-accuracy performance.}
         \label{fig:availability-analysis-time}
     \end{subfigure}
     \hfill
     \begin{subfigure}[t]{0.49\linewidth}
         \centering
         \includegraphics[width=\linewidth]{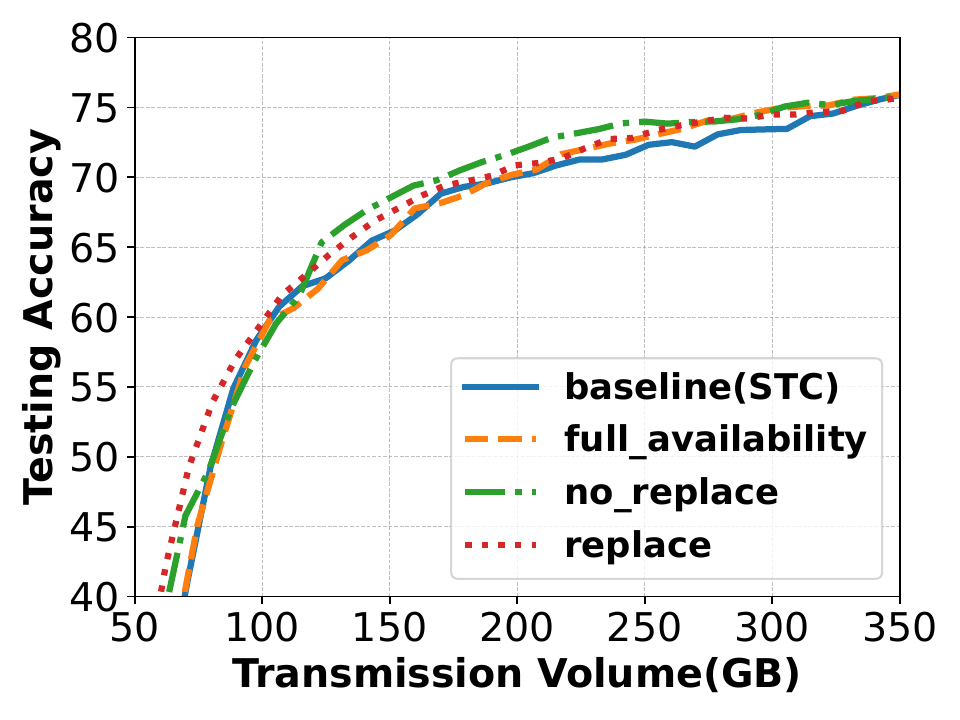}
         \caption{Impact of client availability on bandwidth-to-accuracy performance.}
         \label{fig:availability-analysis-bw}
     \end{subfigure}
    \caption{Impact of Client Availability for \myname + STC.}
    \label{fig:availability-analysis}
\end{figure}

\Cref{fig:availability-analysis} compares two versions of \myname applied to STC. One version is without modifications (\emph{no\_replace}), and the second performs client replacement (\emph{replace}) by including a random client from the set of all clients currently online. We also plot the baseline STC technique under the same availability and FedFetch STC when clients have perfect availability. With clients going offline, the time-to-accuracy performance of \myname diminishes as compared to full availability. However, adding the simple replacement mechanism allows \myname to mitigate most of the effect of offline clients. This is indicated by the \emph{replace} line being further to the left than the \emph{no\_replace} line in \Cref{fig:availability-analysis-time}.

\section{Conclusion}
We introduced \myname, a strategy to address the communication bottleneck in federated learning (FL) caused by the combination of client sampling and update compression techniques. \myname efficiently schedules model state prefetching for clients, significantly reducing download and overall training times. Our evaluation demonstrates that incorporating \myname into communication-efficient FL methods can decrease end-to-end training time by 1.26$\times$ and download time by 4.49$\times$ across a variety of compression techniques. Our code is available at \url{https://github.com/DistributedML/FedFetch}

\bibliographystyle{IEEEtran}
\bibliography{ref}

\end{document}